\def\year{2021}\relax
\documentclass[letterpaper]{article} % DO NOT CHANGE THIS
\usepackage{myaaai21}  
\usepackage{times}  % DO NOT CHANGE THIS
\usepackage{helvet} % DO NOT CHANGE THIS
\usepackage{courier}  % DO NOT CHANGE THIS
\usepackage[hyphens]{url}  % DO NOT CHANGE THIS
\usepackage{graphicx} % DO NOT CHANGE THIS
\usepackage{natbib}  % DO NOT CHANGE THIS AND DO NOT ADD ANY OPTIONS TO IT
\usepackage{caption} % DO NOT CHANGE THIS AND DO NOT ADD ANY OPTIONS TO IT

\nocopyright

\usepackage{times}  % DO NOT CHANGE THIS
\usepackage{helvet} % DO NOT CHANGE THIS
\usepackage{courier}  % DO NOT CHANGE THIS
\usepackage[hyphens]{url}  % DO NOT CHANGE THIS
\usepackage{graphicx} % DO NOT CHANGE THIS
\usepackage{graphicx}  % DO NOT CHANGE THIS
\usepackage{sectsty}

\usepackage{times}
\usepackage{courier}
\usepackage{theorem}
\usepackage{xspace}
\usepackage{boxedminipage}
\usepackage{graphicx}
\usepackage{amssymb}
\usepackage{amsmath}
\usepackage{amsfonts}
\usepackage{alltt}
\usepackage{array}
\usepackage{latexsym}
\usepackage{color}
\usepackage{url}
\usepackage{soul}
\usepackage{tabularx}
\usepackage{amssymb}
\usepackage{tikz}
\usepackage{pgfplots}
\usepackage{bm}
\usepackage{verbatim}
\usepackage{wrapfig}
\usepackage{float}
\usepackage{rotating}
\usepackage{multirow}

\usepackage{algorithm}%,algorithmic}
\usepackage{algpseudocode}
\usepackage{algorithmicx}

\usepackage[toc,page,header]{appendix}
\usepackage{minitoc}

\algdef{SE}[DOWHILE]{Do}{doWhile}{\algorithmicdo}[1]{\algorithmicwhile\ #1}%

\usepackage{mathtools}
\usepackage{amssymb}
\usepackage{amsmath}
\usepackage{xspace}

\newcommand{\reals}{\mathbb{R}}
\newcommand{\tuple}[1]{\langle{#1}\rangle}

\newcommand{\classLabels}{\mathcal{C}}

\newtheorem{example}{Example}[section]

\newtheorem{definition}{Definition}[section]

\def\LTL{\ensuremath{\mathsf{LTL}}\xspace}

\newcommand{\ltluntil}[2]{{#1}\operatorname{\mathcal{U}}{#2}}

\newcommand{\ltlrelease}[2]{{#1}\operatorname{\mathcal{R}}{#2}}
\newcommand{\ltlalways}[1]{\square{#1}}
\newcommand{\tikzcircle}[1]{\tikz[baseline=-0.5ex]\draw[black,radius=3pt,fill=#1] (0,0) circle ;}
\newcommand{\ltlweaknext}[1]{\tikzcircle{black}\xspace{#1}}
\newcommand{\ltlnext}[1]{\tikzcircle{white}\xspace{#1}}
\newcommand{\ltleventually}[1]{\lozenge\xspace{#1}}

\newcommand{\true}{\ensuremath{\mathsf{true}}\xspace}

\usepackage{tikz}
\usetikzlibrary{shapes.geometric}
\usetikzlibrary{positioning,automata,arrows}
\usepackage{pifont}

\usepackage{fontawesome,wasysym,marvosym}

\definecolor{UofTBlue}{RGB}{0,47,101}

\newcommand{\coffee}[0]{{\color{black}\Coffeecup}\xspace}

\usepackage{pgfplots}

\newif\ifcomments
\commentstrue
\ifcomments
\newcommand{\debate}[1]{\textcolor{black}{#1}}

\newcommand{\commentsm}[1]{\textcolor{magenta}{({\bf SM:} #1)}}
\newcommand{\commentms}[1]{\textcolor{magenta}{({\bf MS:} #1)}}

\newcommand{\old}[1]{\textcolor{red}{\sout{#1}}}

\else
 \newcommand{\commentsm}[1]{}
 \newcommand{\commentms}[1]{}
 \newcommand{\old}[1]{}
 
\fi

\newcommand{\revisit}[1]{\textcolor{black}{#1}}

  {\end{alltt}}

\begin{document}

\doparttoc % Tell to minitoc to generate a toc for the parts
\faketableofcontents % Run a fake tableofcontents command for the partocs

% \part{} % Start the document part
% \parttoc % Insert the document TOC

% \newpage

\title{Interpretable Sequence Classification via Discrete Optimization}

\author{Maayan Shvo, Andrew C. Li, Rodrigo Toro Icarte,
Sheila A. McIlraith$^\dagger$ \\}

\affiliations{ 
Department of Computer Science, University of Toronto, Toronto, Canada\\
Vector Institute, Toronto, Canada\\
$^\dagger$~Schwartz Reisman Institute for Technology and Society, Toronto, Canada\\
\{maayanshvo,andrewli,rntoro,sheila\}@cs.toronto.edu}

\maketitle

\setlength\textfloatsep{0.4cm}
\setlength\theorempreskipamount{6pt plus 1pt minus 0.5pt}

\newcommand{\Sys}{\ensuremath{\Sigma}\xspace}
\def\next{\raisebox{1pt}{\begin{footnotesize}\ensuremath{\bigcirc}\end{footnotesize}}}
\newcommand{\eventually}{\begin{large}\ensuremath{\lozenge}\end{large}}
\newcommand{\prefeq}{\ensuremath{\preceq}\xspace}
\newcommand{\pref}{\ensuremath{\prec}\xspace}

\def\noops{\emph{discard}}
\def\noop{\emph{discard}\xspace}

\setcounter{secnumdepth}{2}

\begin{abstract}
Sequence classification is the task of predicting a class label given a sequence of observations.  In many applications such as healthcare monitoring or intrusion detection, early classification is crucial to prompt intervention. In this work, we learn sequence classifiers that favour early classification from an evolving observation trace. While many state-of-the-art sequence classifiers are neural networks, and in particular LSTMs, our classifiers take the form of finite state automata and are learned via discrete optimization. Our automata-based classifiers are interpretable---supporting explanation, counterfactual reasoning, and human-in-the-loop modification---and have strong empirical performance. Experiments over a suite of goal recognition and behaviour classification datasets show our learned automata-based classifiers to have comparable test performance to LSTM-based classifiers, with the added advantage of 
being interpretable.

\end{abstract}
\section{Introduction}

Sequence classification---the task of predicting a class label given a sequence of observations---has a myriad of %real-world 
applications including biological sequence classification (e.g., \cite{deshpande2002evaluation}), document classification (e.g., \cite{sebastiani2002machine}), and intrusion detection (e.g., \cite{lane1999temporal}).
In many settings, early classification is crucial to timely intervention. 
For example, in hospital neonatal intensive care units, early diagnosis of infants with sepsis (based on the classification of sequence data) can be life-saving \cite{griffin2001toward}.

Neural networks such as LSTMs \cite{hochreiter1997long}, learned via gradient descent, are %have been shown to be 
natural and powerful sequence classifiers (e.g., \cite{zhou2015c,karim2019multivariate}), but the rationale for classification can be difficult for a human to discern. This is problematic in many domains, where machine decision-making requires a degree of accountability in the form of verifiable guarantees and explanation for decisions~\cite{doshi2017towards}.

In this work, we use discrete optimization to learn interpretable classifiers that favour early classification. In particular, we learn a suite of binary classifiers that take the form of finite state automata, each capturing the rationale for classification in a compact extractable form. To classify a sequence of observations, we then employ Bayesian inference to produce a posterior probability distribution over the set of class labels. Importantly, our automata-based classifiers, by virtue of their connection to formal language theory, are both generators and recognizers of the pattern language that describes each behavior or sequence class.  We leverage this property in support of a variety of interpretability services, including explanation, counterfactual reasoning, verification of properties, and human modification.

Previous work \debate{on learning automata from data} has focused on learning minimum-sized automata that perfectly classify the training data (e.g., \cite{gold1967language,angluin1987learning,oncina1992identifying,ulyantsev2015bfs,angluin2015learning,giantamidis2016learning,smetsers2018model}). 
\debate{Nonetheless, such approaches learn large, overfitted models in noisy domains that generalize poorly to unseen data. We propose novel forms of regularization to improve robustness to noise and introduce an efficient mixed integer linear programming model to learn these automata-based classifiers. Furthermore, to the best of our knowledge, this is the first work that proposes automata for early classification.}

Experiments on a collection of synthetic and real-world goal recognition and behaviour classification problems demonstrate that our learned classifiers are robust to noisy sequence data, \debate{are well-suited to early prediction}, and achieve comparable performance to an LSTM, with the added advantage of being interpretable.

In Section~\ref{sec:prelim}, we provide necessary background and introduce our running example. In Section~\ref{sec:problem}, we discuss our method for learning DFA sequence classification models and elaborate on the interpretability services afforded by these models in Section~\ref{sec:interp}. 
In Section~\ref{sec:exp}, we discuss the experimental evaluation of our approach on a number of goal recognition and behaviour classification domains, and in Section~\ref{sec:related_work} we situate our work within the body of related work, \revisit{ followed by concluding remarks.}

\section{Background and Running Example}\label{sec:prelim}

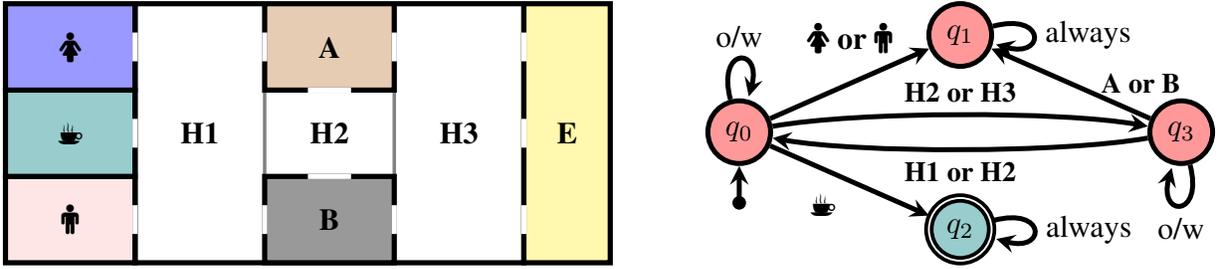
\begin{figure*}[tb]
    \begin{center}
   
    \resizebox{!}{100pt}{\begin{tikzpicture}

    \draw[gray,line width=0.4mm] (6/2,3/3) -- (6/2,6/3);
    \draw[gray,line width=0.4mm] (9/2,3/3) -- (9/2,6/3);

    \draw[draw=black,line width=.6mm,fill=pink!40!white] (0,0) rectangle (3/2,3/3);
    \draw[draw=black,line width=.6mm,fill=teal!40!white] (0,3/3) rectangle (3/2,6/3);
    \draw[draw=black,line width=.6mm,fill=blue!40!white] (0,6/3) rectangle (3/2,9/3);
    \draw[draw=black,line width=.6mm] (3/2,0) rectangle (12/2,9/3);
    \draw[draw=black,line width=.6mm,fill=black!40!white] (6/2,0) rectangle (9/2,3/3);
    \draw[draw=black,line width=.6mm,fill=brown!40!white] (6/2,6/3) rectangle (9/2,9/3);
    \draw[draw=black,line width=.6mm,fill=yellow!40!white] (12/2,0) rectangle (14/2,9/3);
    
    % doors
    \draw[white, line width=.6mm] (3/2,1/3) -- (3/2,2/3);
    \draw[white, line width=.6mm] (3/2,4/3) -- (3/2,5/3);
    \draw[white, line width=.6mm] (3/2,7/3) -- (3/2,8/3);
    
    \draw[white, line width=.6mm] (6/2,1/3) -- (6/2,2/3);
    \draw[white, line width=.6mm] (6/2,7/3) -- (6/2,8/3);
    
    \draw[white, line width=.6mm] (9/2,1/3) -- (9/2,2/3);
    \draw[white, line width=.6mm] (9/2,7/3) -- (9/2,8/3);
    
    \draw[white, line width=.6mm] (12/2,1/3) -- (12/2,2/3);
    \draw[white, line width=.6mm] (12/2,4/3) -- (12/2,5/3);
    \draw[white, line width=.6mm] (12/2,7/3) -- (12/2,8/3);
    
    \draw[white, line width=.6mm] (7/2,3/3) -- (8/2,3/3);
    \draw[white, line width=.6mm] (7/2,6/3) -- (8/2,6/3);

    % objects
    \node at (7.5/2,7.5/3) {\textbf{A}};
    \node at (7.5/2,4.5/3) {\textbf{H2}};
    \node at (7.5/2,1.5/3) {\textbf{B}};
    
    \node at (4.5/2,4.5/3) {\textbf{H1}};
    \node at (10.5/2,4.5/3) {\textbf{H3}};
    
    \node at (1.5/2,7.5/3) {\textbf{\faFemale}};
    \node at (1.5/2,4.5/3) {\textbf{\coffee}};
    \node at (1.5/2,1.5/3) {\textbf{\faMale}};
    
    \node at (13/2,4.5/3) {\textbf{E}};

\end{tikzpicture}%
}
    \hspace*{3em}
    \resizebox{!}{100pt}{\begin{tikzpicture}[node distance=2cm,on grid,every initial by arrow/.style={ultra thick,->, >=stealth}, initial text={}]%
  \node[ultra thick,state,fill=red!40!white,minimum size=.5cm] (q_0) at (-3/2,0) {$q_0$};
  \node[ultra thick,state,fill=red!40!white,minimum size=.5cm]  (q_3) at (7/2,0) {$q_3$};
  \node[very thick,state,accepting, fill=teal!40!white,minimum size=.5cm]  (q_2) at (2/2,-1.1) {$q_2$};
  \node[ultra thick,state,fill=red!40!white,minimum size=.5cm]  (q_1) at (2/2,1.1) {$q_1$};

  % Initial node
  \path[ultra thick,->, >=stealth] (-3/2,-0.8) edge node [right] {} (q_0);
  \draw[fill=black] (-3/2,-0.8) circle (0.07);
  
  % Loops
  \path[->] (q_0) edge [ultra thick,->, >=stealth,loop above] node { \text{o/w}} ();
  \path[->] (q_1) edge [ultra thick,->, >=stealth,loop right] node { \text{always}} ();
  \path[->] (q_2) edge [ultra thick,->, >=stealth,loop right] node { \text{always}} ();
  \path[->] (q_3) edge [ultra thick,->, >=stealth,loop below] node { \text{o/w}} ();
  
  % Edges
  \path[ultra thick,->, >=stealth] (q_0) edge node [above=0.2] { {\faFemale} \textbf{or} {\faMale}} (q_1);
  
  \path[ultra thick,->, >=stealth] (q_0) edge node [below left] {{\coffee}} (q_2);
  
  \draw[ultra thick,->, >=stealth, out=10, in=170, looseness=0.6] (q_0) to node [above] {\textbf{\small H2 or H3}} (q_3);
  
  \draw[ultra thick,->, >=stealth, out=190, in=-10, looseness=0.6] (q_3) edge node [below] {\textbf{\small H1 or H2}} (q_0);
  
  \path[ultra thick,->, >=stealth] (q_3) edge node [right=0.2] {\textbf{\small A or B}} (q_1);

\end{tikzpicture}%
}
    
    \caption{\textbf{Left} - Goal recognition environment where the possible goals of the agent are going
    to an office %work
    (\textbf{A} or \textbf{B}), leaving the building (\textbf{E}), going to the restroom (\faFemale $\ $or \faMale), or getting coffee (\coffee). \textbf{Right} - a DFA classifier that detects whether or not the agent is trying to reach the goal {\coffee}.  
    A decision is provided after each new observation based on the current state: \textbf{yes} for the blue accepting state, and \textbf{no} for the red, non-accepting states. ``o/w" (otherwise) stands for all symbols that do not appear on outgoing edges from a state. ``always" stands for all symbols. The DFA is guaranteed to correctly classify traces from an agent starting in \textbf{A}, \textbf{B}, or \textbf{E} that pursues an optimal path using only the hallways, measured by Manhattan distance. It also learns to generalize to some 
    traces not seen in training. E.g., the trace (\textbf{B}, \textbf{H3}, \textbf{H2}, \textbf{H1}, \coffee) is accepted and (\textbf{B}, \textbf{H2}, \textbf{H1}, \faMale) is rejected.
    }
    
    \label{fig:running_example}
    \end{center}

\end{figure*}

The class of problems we address are symbolic time-series classification problems that require discrimination of a set of potential classes, where early classification may be favored, data may be noisy, and an interpretable, and ideally queryable, classifier is either necessary or desirable.

We define the sequence classification problem as follows.

\begin{definition}[Sequence Classification]
Given a trace $\tau = (\sigma_1, \sigma_2, \ldots, \sigma_n)$, $\sigma_i \in \Sigma$, where $\Sigma$ is a finite set of symbols, and $\classLabels$ is a set of class labels, sequence classification is the task of predicting the class label $c \in \classLabels$ that corresponds to $\tau$.
\end{definition}

The observation trace, $\tau$, is typically assumed to encode the entire trace. However, we also examine the \emph{early classification} setting, where it is desirable to produce a high confidence classification with a small prefix of the entire trace (e.g., \cite{griffin2001toward,xing2010brief,ghalwash2013extraction}).

 We propose the use of Deterministic Finite Automata (DFA) as sequence classifiers.

\begin{definition}[Deterministic Finite Automaton]
A Deterministic Finite Automaton is a tuple $\mathcal{M}$ $=$ $\tuple{Q,q_0,\Sigma,\delta,F}$, where $Q$ is a finite set of states, $q_0 \in Q$ is the initial state, $\Sigma$ is a finite set of symbols, $\delta: Q \times \Sigma \to Q$ is the state-transition function, and $F \subseteq Q$ is a set of accepting states.  
\end{definition}

Given a sequence of input symbols $\tau = (\sigma_1, \sigma_2, \ldots, \sigma_n)$, $\sigma_i \in \Sigma$, a DFA $\mathcal{M}=\tuple{Q,q_0,\Sigma,\delta,F}$ transitions through the sequence of states $s_0, s_1, \ldots, s_n$ where $s_0 = q_0$, $s_i = \delta(q_{i-1}, \sigma_i)$ for all $1 \le i \le n$. $\mathcal{M}$ \emph{accepts} $\tau$ if $s_n \in F$, otherwise, $\mathcal{M}$ \emph{rejects} $\tau$.

A DFA provides a compact graphical encoding of a \emph{language}, a set of (potentially infinite) traces accepted by the DFA. The class of languages recognized by DFAs is known collectively as the \emph{regular languages}. 
In Section~\ref{sec:interp} we employ formal language theory to straightforwardly propose a set of interpretability services over our DFA classifiers.

We use the following goal recognition problem as a running example to help illustrate concepts.

\begin{example}[The office domain]
\label{running_example}
Consider the environment shown in Figure~\ref{fig:running_example}. We observe an agent that starts at one of %the locations 
\textbf{A}, \textbf{B}, or \textbf{E} with the goal of reaching one of the other coloured regions,  %areas in the diagram 
$\classLabels = \{\text{\textbf{A}},\text{\textbf{B}},\text{\textbf{E}}, \text{\coffee}, \text{\faFemale}, \text{\faMale}\}$, using \emph{only} the hallways \textbf{H1}, \textbf{H2}, and \textbf{H3}. The agent always takes the shortest Manhattan distance path to the goal,  %(by Manhattan distance), 
choosing uniformly at random if multiple shortest paths exist. E.g., an agent starting at \textbf{B} with goal {\coffee} will pursue paths (\textbf{B, H2, H1} {\coffee})  and (\textbf{B, H1,} {\coffee}). We wish to predict the agent's goal as early as possible, given a sequence of observed locations.

\end{example}

Figure~\ref{fig:running_example} shows a binary DFA classifier that predicts if an agent is trying to achieve the \coffee goal in the office domain. This DFA was learned from the set of all \emph{valid} traces and corresponding goals (as stipulated in Example~\ref{running_example}) using the method discussed in Section~\ref{sec:problem}. 
The DFA predicts whether the agent would achieve the \coffee goal by keeping track of the agent's locations over time. Its input symbols are $\Sigma = \{$\textbf{A}, \textbf{B}, \textbf{H1}, \textbf{H2}, \textbf{H3}, \textbf{E}, \faFemale, \faMale, \coffee$\}$ and the only accepting state is $q_2 \in F$.
A decision is provided after each incoming observation based on the current state: \textbf{yes} for the blue accepting state, and \textbf{no} for red, non-accepting states. For example, on the trace (\textbf{B}, \textbf{H2}, \textbf{H1}, \coffee) the DFA would transition through the states $(q_0, q_3, q_0, q_2)$, predicting that the goal is \emph{not} \coffee after the first three observations, then predicting the goal \emph{is} \coffee after the fourth observation.

Note that this learned DFA leverages biases in the data---namely, that in the training data the agent only pursues optimal paths. However, in addition to correctly classifying all optimal paths, the DFA also generalizes to some unseen traces.
For example, the DFA correctly classifies the trace corresponding to the suboptimal path (\textbf{B}, \textbf{H3}, \textbf{H2}, \textbf{H1}, \coffee).

Finally, the DFA in Figure~\ref{fig:running_example} only predicts the goal \coffee once \coffee is observed. Another DFA trained to detect goal \textbf{E} (not shown) highlights early detection by predicting goal \textbf{E} once \textbf{H3} is observed, unless the agent started at \textbf{E}. This is correct since when the agent starts at \textbf{A} or \textbf{B}, the observation \textbf{H3} only appears on optimal paths to \textbf{E}.

\section{Learning DFAs for Sequence Classification}
\label{sec:problem}

In this section, we describe our method for learning DFA sequence classification models from a set of training traces and corresponding class labels $\{(\tau_1, c_1), \ldots, (\tau_N, c_N)\}$. We adopt the standard supervised learning assumption that each $(\tau_i, c_i) \stackrel{iid}{\sim} p(\tau, c)$, and the objective of maximizing the predictive accuracy of the model over $p(\tau, c)$. For each possible label $c \in \classLabels$, we train a separate DFA $\mathcal{M}_c$ responsible for recognizing traces with label $c$ (see Section~\ref{sec:ovr}). At test time, given a trace (or a partial trace) $\tau$, all $|\classLabels|$ DFAs are evaluated independently on $\tau$, and the collective decisions of the DFAs are used to produce a posterior probability distribution over $\classLabels$ (see Section~\ref{sec:bayes}).

\subsection{Learning one-vs-rest binary classifiers}
\label{sec:ovr}

We now describe our \emph{Mixed Integer Linear Programming (MILP)} model to learn DFAs. We rely on MILP solvers because they are the state of the art for solving a wide range of discrete optimization problems and they are guaranteed to find optimal solutions given sufficient resources \cite{junger200950}.

Given a training set $\{(\tau_1, c_1), \ldots, (\tau_N, c_N)\}$, we learn one DFA $\mathcal{M}_c$ per each label $c \in \classLabels$ responsible for discriminating traces in class $c$ from traces not in class $c$. 

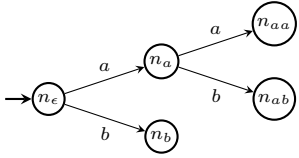
\begin{figure}
    \centering
    \begin{tikzpicture}[on grid,initial distance=10pt,
  every initial by arrow/.style={thick,->, >=stealth}, initial text={}]%
  \node[thick,state,initial,inner sep=1pt,minimum size=0pt] (u_0) at (0,0) {\scriptsize$n_{\epsilon}$};
  \node[thick,state,inner sep=1pt,minimum size=0pt] (u_1) at (1.5,.5) {\scriptsize$n_{a}$};
  \node[thick,state,inner sep=1pt,minimum size=0pt] (u_2) at (1.5,-.5) {\scriptsize$n_{b}$};
  \node[thick,state,inner sep=1pt,minimum size=0pt] (u_3) at (3,1) {\scriptsize$n_{aa}$};
  \node[thick,state,inner sep=1pt,minimum size=0pt] (u_4) at (3,0) {\scriptsize$n_{ab}$};

   \path[->, >=stealth] (u_0) edge node [above]{\scriptsize$a$} (u_1);
   \path[->, >=stealth] (u_0) edge node [below]{\scriptsize$b$} (u_2);
   \path[->, >=stealth] (u_1) edge node [above]{\scriptsize$a$} (u_3);
   \path[->, >=stealth] (u_1) edge node [below]{\scriptsize$b$} (u_4);

\end{tikzpicture}%
    \caption{A PT for $\{b,aa,ab\}$.} %\cite{giantamidis2016learning}}
    \label{fig:pt_example}
\end{figure}

We start by representing the whole training set as a \emph{Prefix Tree (PT)} \cite{de2010grammatical}---this is a common preprocessing step in the automata learning literature \cite{giantamidis2016learning}. PTs are incomplete DFAs with no accepting states. They are incomplete in the sense that some of their transitions are unspecified. Given a training set $\{(\tau_1, c_1), \ldots, (\tau_N, c_N)\}$, we can construct (in polynomial time) a PT that compactly represents all the prefixes from the training traces. In particular, the PT will be $\mathcal{P} = \tuple{N,n_{\epsilon},\Sigma,\delta,\emptyset}$ such that $\Sigma$ is the set of symbols in the training set and, for every training trace $\tau_i=(\sigma_1,\ldots,\sigma_n)$, 1) there is a node $n_{(\sigma_1,\ldots,\sigma_j)} \in N$ for all $j\in\{0\ldots n\}$ (i.e., all prefixes of $\tau_i$) and 2) the transition function is constrained to have $\delta(n_{(\sigma_1,\ldots,\sigma_j)}, \sigma_{j+1}) = n_{(\sigma_1,\ldots,\sigma_{j+1})}$ for all $j\in \{2,n-1\}$ and $\delta(n_{\epsilon},\sigma_1) = n_{\sigma_1}$. Intuitively, this PT defines a tree where all training traces are branches. As an example, Figure~\ref{fig:pt_example} shows the PT for a set of training traces $\{b,aa,ab\}$ (from \citet{giantamidis2016learning}).

Each node in the PT represents a prefix that appears in one or more training traces. After constructing the PT, we label its nodes with the number of positive $n^+$ and negative $n^-$ training traces that start with the node's prefix. Positive traces are those belonging to the target class $c$ for learning DFA $\mathcal{M}_c$ and all other traces are negative traces. In the example, if traces $b$ and $aa$ are positive and $ab$ is negative, then $n_{\epsilon}^+=2$, $n_{\epsilon}^-=1$, $n_{b}^+=1$, $n_{b}^-=0$, $n_{a}^+=1$, $n_{a}^-=1$, $n_{aa}^+=1$, $n_{aa}^-=0$, $n_{ab}^+=0$, and $n_{ab}^-=1$. In practice, we divide the contribution of each $\tau_i$ to $n^+$ or $n^-$ by its length so that longer traces are not overrepresented. We then use these values to compute the training error in our MILP model.

Our MILP model takes the PT $\mathcal{P} = \tuple{N,n_{\epsilon},\Sigma,\delta,\emptyset}$, its nodes counters ($n^+$ and $n^-$), and a positive number $q_{\max}$ to learn a DFA $\mathcal{M}_c=\tuple{Q,q_0,\Sigma,F}$ for class $c$ with at most $q_{\max}$ states.
The main idea is to assign the DFA state reached by each node in the tree (which represents a sequence of observations). A binary decision variable $x_{nq}$, which is equal to 1 if DFA state $q$ is assigned to node $n$ (and zero otherwise) encodes this assignment based on the transition function $\delta$. Our model searches for an assignment of DFA states to the tree nodes that is feasible (there exists a transition function $\delta$ that generates such an assignment) and has low early prediction error. We predefine a set of accepting states $F$ and add the error $x_{nq}n^-$ if $q \in F$ and by $x_{nq}n^+$ if $q \not\in F$, for all $n\in N$. 

To reduce overfitting, we limit the maximum number of DFA states, a common form of regularization in automata learning \cite{gold1967language,gold1978complexity,giantamidis2016learning}. Additionally, we designate two DFA states---one accepting, and one non-accepting---as \emph{absorbing states} which can only self-transition. These states prevent the classifier from changing decisions once reached. We found that rewarding nodes for reaching these absorbing states and penalizing the number of transitions between different DFA states acted as effective regularizers, significantly improving generalization. Further details about the MILP model can be found in the Technical Appendix \S~A. 
%~\ref{sup:learn_dfa}.

In principle, the binary DFA classifiers learned by the MILP model can be used directly. However, they would have deterministic outcomes (i.e., the trace is either accepted or rejected), might contradict each other (multiple DFAs could accept the same trace), or might all reject the trace. In the next section, we address these issues by computing a probability distribution from the DFAs' predictions.

\subsection{Posterior inference of the class label}
\label{sec:bayes}

Given a trace (or partial trace) $\tau$ and the decisions of the one-vs-rest classifiers $\{\mathrm{D}_c(\tau) : c \in \classLabels \}$, we use an approximate Bayesian method to infer a posterior probability distribution over the true label $c^{*}$. Each $D_c(\tau)$ is treated as a discrete random variable with possible outcomes $\{\mathrm{accept}, \mathrm{reject}\}$. We make the following assumptions: {(1) the classification decisions $\mathrm{D}_c$ for $c \in \classLabels$ are conditionally independent given the true label $c^{*}$} and {(2) $p(D_c | c^*)$ only depends on whether $c = c^*$.

For each $c'$, we compute the posterior probability of $c^* = c'$ to be

\begin{align*}
    \hspace{5mm} p(c^* = c' | \{ \mathrm{D}_c : c \in \classLabels \} ) \propto \frac{p(c^* = c') * p(D_{c'} | c^* = c')} { p(D_{c'} | c^* \neq c') }
\end{align*}
\noindent The full derivation can be found in the Technical Appendix \S~A.3. The probabilities on the right-hand side are estimated using a held-out validation set. 
% In particular, the prior probability $p(c^* = c')$ is the approximate proportion of examples with label $c'$, $p(D_{c'} = \mathrm{accept} | c^* = c')$ is the recall of the classifier for label $c'$, and so on. 
We normalize the posterior probabilities $p(c^* = c' | \{ \mathrm{D}_c : c \in \classLabels \} )$ such that their sum over $c' \in \classLabels$ is 1 to obtain a valid probability distribution. Note that we could potentially improve the inference by further conditioning on the number of observations seen so far, or relaxing assumption (2). However, this would substantially increase the number of probabilities to be estimated and result in less accurate estimates in low-data settings.

\subsection{Discussion}

In this section, we described an approach to learning DFAs for sequence classification based on mixed integer linear programming. Our model includes a set of constraints to enforce DFA structure, but in general, constraints can also be added
to incorporate domain knowledge. Furthermore, our formulation is compatible with off-the-shelf optimizers and as such can benefit from advances in discrete optimization solvers. While our approach does not scale as well as gradient-based optimization, our use of prefix trees significantly reduces the size of the discrete optimization problem, allowing us to \revisit{tackle} real-world datasets with nearly 100,000 observation tokens, as demonstrated in our experiments.

We further show how the decisions of the DFA classifiers can be used to predict a probability distribution over labels. In our experiments we demonstrate how these probabilities are useful in an early classification setting. Furthermore, we can flexibly handle many important settings in sequence classification, including returning the $k$ most probable class labels and multi-label sequence classification (see Technical Appendix \S~C.5).

\section{Classifier Interpretability}
\label{sec:interp}

An important property of our learned classifiers is that they are interpretable insofar as that the rationale leading to a classification is captured explicitly in the structure of the DFA.
%directly or via one of the services described below. Indeed
DFAs can be queried and manipulated to provide a set of interpretability services including explanation, verification  of classifier properties, and (human) modification, as we demonstrate below. 
Our purpose here is to highlight the breadth of interpretability services afforded by DFA classifiers via their relationship to formal language theory. The effectiveness of a particular interpretability service is user-, domain-, and even task-specific and is best evaluated in the context of individual domains. We leave detailed study of this question to a separate paper.

As noted in Section \ref{sec:prelim}, DFAs provide a compact, graphical representation of a (potentially infinite) set of traces the DFA positively classifies. Collectively, each DFA defines a regular language, the simplest form of language in the Chomsky hierarchy \cite{chomsky1956three}. While many people will find the DFA structure highly interpretable, the DFA classifier can be transformed into a variety of different language-preserving representations including regular expressions, context-free grammars (CFGs), and variants of Linear Temporal Logic (LTL)~\cite{pnueli1977temporal} (see also Technical Appendix \S~B.3). These transformations are automatic and can be decorated with natural language to further enhance human interpretation.

\begin{example}
The following regular expression compactly describes the set of traces that are classified as belonging to the DFA classifier depicted in Figure \ref{fig:running_example}:
$[(\Sigma - \{ \text\faFemale, \text\faMale, H2, H3 \})^*(H2 \cup H3) (\Sigma - \{A,B,H1,H2\})^*(H1 \cup H2)]^*(\Sigma - \{\text\faFemale, \text\faMale, H2,H3 \})^* \text\coffee \Sigma^*$
\end{example}
Of course, this regular expression \revisit{is only decipherable to a subset of computer scientists.} \revisit{We include it in order to illustrate/demonstrate the multiple avenues for interpretation afforded by our DFA classifiers}. \revisit{In particular, }the regular expression can be further transformed into a more human-readable form as illustrated in Example~\ref{ex:re2english} or transformed into a CFG that is augmented with natural language in order to provide an enumeration, or if abstracted, a compact description of the traces accepted by the DFA classifier. 

\begin{example}\label{ex:re2english}The regular expression can be transformed into a more readable form such as: \\
"Without first doing  \text\faFemale ~or~ \text\faMale, repeat the following zero or more times: eventually do H2 or H3, then without doing  A or B, eventually do H1 or H2, followed optionally by other events, excluding \faFemale ~and~ \faMale.
 Finally do \coffee, followed by anything."
 \end{example}

For others, it may be  informative to extract path properties of a DFA as LTL formulae, perhaps over a subset of $\Sigma$ or with preference for particular syntactic structures (e.g., \cite{camacho2019learning}).

\begin{example}
DFA classifier ${\cal M} \models  \ltlalways{\ltleventually{\text{\coffee}}}$, the LTL property "always eventually do \coffee".

\end{example}

These transformations and entailments utilize well studied techniques from formal language theory (e.g., \cite{rozenberg2012handbook}). Which delivery form is most effective is generally user- and/or task-specific and should be evaluated in situ via a usability study.

\subsection{Explanation}

An important service in support of interpretability is explanation. In the context of classification, given classifier $\mathcal{M}$ and trace $\tau$, we wish to query $\mathcal{M}$, seeking explanation for the classification of $\tau$. 

In many real-world applications, traces comprise extraneous symbols that are of no interest %to the human 
and play no role in the classifier (such as the agent scratching their nose en route to %getting 
coffee). It often makes sense to define an \emph{explanation vocabulary}, $\Sigma_{e} \subseteq \Sigma$, a set of distinguished symbols of interest for explaining classifications that are pertinent to the explanation of traces such as $\tau$, i.e. $\Sigma_{e} \cap \Sigma_{\tau} \ne \{\}$. 
Explanations for a positive classification can be extracted from a DFA classifier over an explanation vocabulary following the techniques described above.

In cases where a classifier does not return a positive classification for a trace, a useful explanation can take the form of a so-called \emph{counterfactual explanation} (e.g., \cite{miller2019explanation}).

Let $\alpha$ and $\beta$ be strings over $\Sigma$. The \textit{edit distance} between $\alpha$ and $\beta$, $d(\alpha, \beta)$, is equal to the minimum number of edit operations required to transform $\alpha$ to $\beta$. We take the edit distance between two strings to be their Levenshtein distance where the set of edit operations comprises \textit{insertion}, \textit{deletion}, and \textit{substitution}, and where each of these operations has unit cost \cite{levenshtein1966binary}.

\begin{definition}[Counterfactual Explanation]\label{def:counterfactual_exp}

Let $\mathcal{M}$ be a DFA classifier that accepts the regular language $\mathcal{L}$ defined over $\Sigma$ and let $\tau$ be some string over $\Sigma$. A counterfactual explanation for $\tau$ is the sequence of edit operations transforming $\tau$ to a string $\tau' = argmin_{\omega \in \mathcal{L}}( { d(\tau, \omega) })$.

\end{definition}

\citeauthor{wagner1974order} (\citeyear{wagner1974order}) proposed an algorithm that computes, for some string $\tau$ and regular language $\mathcal{L}$, a string $\tau' \in \mathcal{L}$ with minimal edit distance from $\tau$. The algorithm has a time complexity that is quadratic in the number of states of the DFA that accepts the language $\mathcal{L}$ in question.

\begin{example}
Given the DFA depicted in Figure~\ref{fig:running_example} and a trace $\tau = $ (\textbf{A, H2, H1,} {\faMale}), a possible counterfactual explanation is the edit operation (informally specified) \textsc{Replace} \faMale\hspace{0.1cm}\textsc{with} \coffee which transforms (\textbf{A, H2, H1,} {\faMale}) to (\textbf{A, H2, H1,} {\coffee}).
This explanation can then be transformed into a natural language sentence: \textit{``The binary classifier would have accepted the trace had \coffee been observed instead of \faMale"}.
A simple approach that generates such natural language sentences from counterfactual explanations can be found in the Technical Appendix \S~B.1.
\end{example}

\subsection{Classifier Verification and Modification}

Explanation encourages human trust in a classification system, but it can also expose rationale that prompts a human (or automated system) to further question or to wish to modify the classifier. Temporal properties of the DFA classifier ${\cal M}$, such as \emph{``Neither \faFemale~nor~\faMale~ occur before \coffee''} can be straightforwardly specified in LTL and verified against ${\cal M}$ using standard %sound and complete
formal methods verification techniques (e.g., \cite{vardi1986automata}). In the case where the property is false, a witness can be be returned.

Our learned classifiers are also amenable to the inclusion of additional classification criteria, and the modification to the DFA classifier can be realized via a standard product computation. 

Let $\mathcal{L}_1$ and $\mathcal{L}_2$ be regular languages over $\Sigma$. Their intersection is defined as $\mathcal{L}_1 \cap \mathcal{L}_2 = \{x \mid x \in \mathcal{L}_1 \text{ and } x \in \mathcal{L}_2 \}$. Let $\mathcal{M}_1$ and $\mathcal{M}_2$ be the DFAs that accept $\mathcal{L}_1$ and $\mathcal{L}_2$, respectively. The \textit{product} of $\mathcal{M}_1$ and $\mathcal{M}_2$ is $\mathcal{M}_1 \times \mathcal{M}_2$ where the language accepted by the DFA $\mathcal{M}_1 \times \mathcal{M}_2$ is equal to $\mathcal{L}_1 \cap \mathcal{L}_2$ (i.e., $\mathcal{L}(\mathcal{M}_1 \times \mathcal{M}_2) = \mathcal{L}_1 \cap \mathcal{L}_2$).

\begin{definition}[Classifier Modification]
Given a DFA encoding some classification criterion $\mathcal{M}_c$ and a DFA classifier $\mathcal{M}$, the modified classifier $\mathcal{M}'$ is the product of $\mathcal{M}$ and $\mathcal{M}_c$. %$\mathcal{M} \times \mathcal{M}_c$.
\end{definition}

Classifier modification ensures the enforcement of criterion ${\cal M}_c$ in ${\cal M}'$. However, such post-training modification could result in rejection of traces in the dataset that are labelled as positive examples of the class. Such modification can (and should) be verified against the dataset.
Finally, modification criteria can be expressed directly in a DFA, or specified in a more natural form such as LTL.

\section{Experimental Evaluation}
\label{sec:exp}

In this section we evaluate our approach, Discrete Optimization for Interpretable Sequence Classification (DISC), on a suite of goal recognition and behaviour classification domains. \revisit{DISC is the implementation of the MILP model and Bayesian inference method described in Section~\ref{sec:problem}.} We compare against LSTM \cite{hochreiter1997long}, a state-of-the-art neural network architecture for sequence classification; Hidden Markov Model (HMM), a probabilistic generative model which has been extensively applied to sequence tasks \cite{kupiec1992robust,sonnhammer1998hidden}; n-gram \cite{dunning1994statistical} for $n=1,2$, which perform inference under the simplifying assumption that each observation only depends on the last $n-1$ observations; and a DFA-learning approach (DFA-FT) which maximizes training accuracy (minimizing the number of DFA states only as a secondary objective), representative of existing work in learning DFAs.

DISC, DFA-FT, and HMM learn a separate model for each label while LSTM and $n$-gram directly model a probability distribution over labels. Each classifier predicts the label with highest probability and all datasets consist of at least $7$ labels. The LSTM optimized average accuracy over all prefixes of an observation trace in order to encourage early prediction.

Table \ref{tab:summary} contains a summary of results for all datasets.
Additional results (including examples of DFA classifiers learned from the data) and details of the experiments can be found in the Technical Appendix (\S~B.2 and \S~C.3). The code for DISC is available online\footnote{https://github.com/andrewli77/DISC}.

\subsection{Experimental Setup}

Performance is measured as follows. Cumulative convergence accuracy (CCA) at time $t$ is defined as the percentage of traces $\tau$ that are correctly classified given $\mathrm{min}(t, |\tau|)$ observations. Percentage convergence accuracy (PCA) at $X$\% is defined as the percentage of traces $\tau$ that are correctly classified given the first $X$\% of observations from $\tau$. All results are averaged over 30 runs, unless otherwise specified.

\debate{The datasets we used for evaluation were selected to be representative of the diversity of this class, both with respect to data properties such as noise, complexity of classification patterns, and to be somewhat suggestive of the diversity of the tasks for which this work is applicable. }
We considered three goal recognition domains: Crystal Island \cite{ha2011goal,min2016player}, a narrative-based game where players solve a science mystery; ALFRED \cite{ALFRED20}, a virtual-home environment where an agent can interact with various household items and perform a myriad of tasks; and MIT Activity Recognition (MIT-AR) \cite{tapia2004activity}, comprised of noisy, real-world sensor data with labelled activities in a home setting. Given a trace the classifier attempts to predict the goal the agent is pursuing. Crystal Island and MIT-AR are particularly challenging as subjects may pursue goals non-deterministically.

Experiments for behaviour classification were conducted on a dataset comprising replays of different types of scripted agents in the real-time strategy game StarCraft \cite{kantharajuMCTS2019}, and on two real-world malware datasets comprising `actions' taken by different malware applications in response to various Android system events (BootCompleted and BatteryLow) \cite{bernardi2019dynamic}.
The behaviour classification task involves predicting the type of StarCraft agent and malware family, respectively, that generated a given behaviour trace.

\subsection{Results}
\label{sec:results}

\renewcommand{\arraystretch}{0.93}

  \setlength{\tabcolsep}{2.7pt} 

\newcommand{\tb}[1]{\textbf{#1}}

\begin{table*}
\centering

\begin{tabular}{l|ll|p{1.5cm}p{1.5cm}p{1.5cm}p{1.5cm}p{1.5cm}p{1.5cm}}
                        & & & \multicolumn{6}{c}{Percent Accuracy given full observation traces} \\ \hline
Dataset                  & $N$    & $|\tau|$ & \textbf{DISC} & DFA-FT &  LSTM & HMM & 1-gram & 2-gram \\ \hline

Crystal Island          & 893   & 52.9  & 78 ($\pm 1.2$)    & 46 ($\pm 1.0$)    & \tb{87} ($\pm 1.1$)   & 57 ($\pm 1.2$)    & 69 ($\pm 0.9$)    & 57 ($\pm 1.1$)    \\
StarCraft               & 3872  & 14.8  & 43 ($\pm 0.4$)     & 38 ($\pm 0.4$)    & \tb{44} ($\pm 0.4$)    & 38 ($\pm 0.6$)   & 29 ($\pm 0.4$)   & 37 ($\pm 0.4$)    \\
ALFRED                  & 2520  & 7.5   & \tb{99} ($\pm 0.1$)    & 94 ($\pm 0.3$)    & \tb{99} ($\pm 0.1$)    & 97 ($\pm 0.7$)    &  83 ($\pm 0.3$)    & 94 ($\pm 0.2$)    \\
MIT-AR                  & 283   & 9.3   & 57 ($\pm 1.9$)   &  36 ($\pm 1.9$)    & 56 ($\pm 1.9$)    & 45 ($\pm 2.1$)    & \tb{66} ($\pm 2.0$)    & 55 ($\pm 1.5$)    \\
BootCompleted           & 477   & 206.0 & 59 ($\pm 2.2$)    & \tb{69} ($\pm 1.5$)   &  65 ($\pm 1.3$)   & 54 ($\pm 2.4$)    & 46 ($\pm 2.8$)    & 55 ($\pm 1.6$)    \\
BatteryLow              & 283   & 216.2 & 60 ($\pm 1.4$)    & \tb{73} ($\pm 1.4$)    &  70 ($\pm 1.4$)    &  52 ($\pm 2.2$)   & 35 ($\pm 1.3$)   & 54 ($\pm 1.5$)   \\
\end{tabular}

\caption{A summary of results from all domains (\textbf{DISC} is our approach). With respect to the full dataset, $N$ is the total number of traces, and $|\tau|$ is the average length of a trace. Reported are the percentages of traces correctly classified given the full observation trace, with 90\% confidence error in parentheses. Highest accuracy is bolded. }
\label{tab:summary}
\end{table*}

\begin{figure*}[tb]
    \centering
    %{ StarCraft-a \qquad ~~~~~~~~~~~~~~~~~~~~~~~~~ BC $\qquad$}
    %\vspace{1mm}
        
    \includegraphics[width=\textwidth]{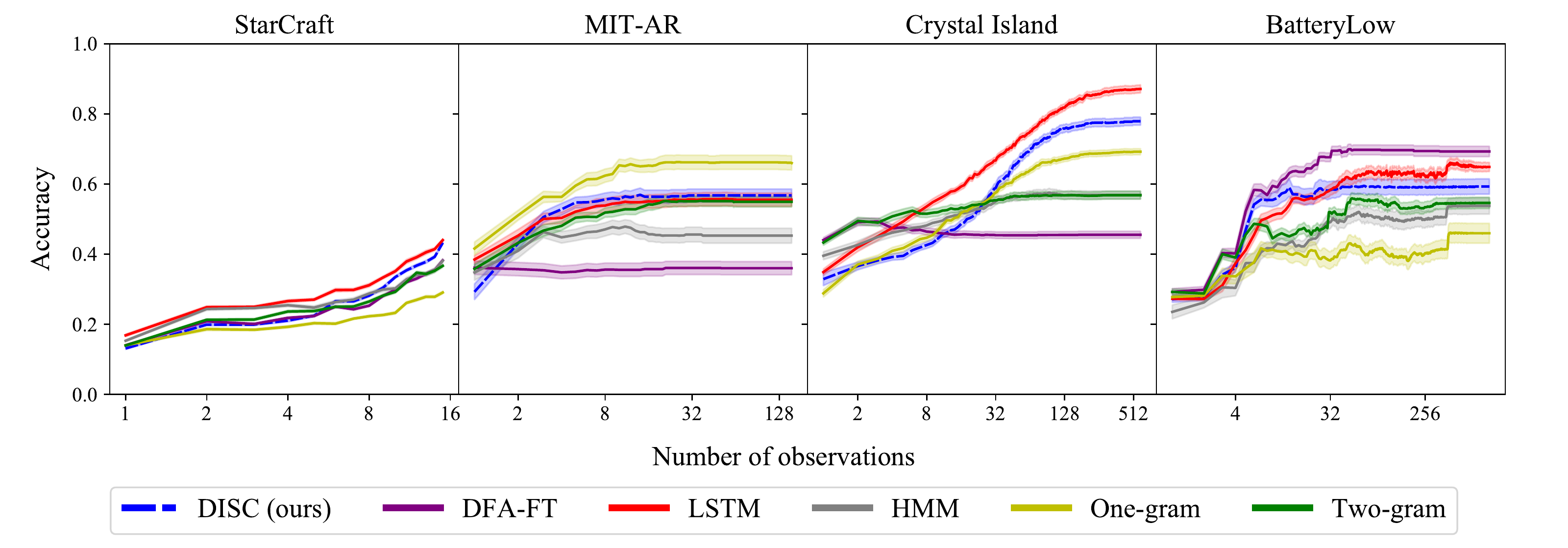}
    \hfill

    \vspace{-8mm}
    \caption{Test accuracy of DISC and all baselines as a function of earliness (number of observations seen so far)
    on one synthetic dataset (left) and three real-world datasets (three right). We report %CCA 
    \debate{Cumulative Convergence Accuracy} up to the maximum length of a trace. Error bars correspond to a 90\% confidence interval. \revisit{Further results appear in the Technical Appendix \S~C.}}
    \label{fig:results}
\end{figure*}

Detailed results for StarCraft, MIT-AR, Crystal Island, and BatteryLow are shown in Figure \ref{fig:results}
while a summary of results from all domains is provided in Table \ref{tab:summary}.

DISC generally outperformed n-gram, HMM, and DFA-FT, achieving near-LSTM performance on most domains. LSTM displayed an advantage over DISC on datasets with long traces. n-gram models excelled in some low-data settings (see MIT-AR) but perform poorly overall as they fail to model long-term dependencies. Surprisingly, DFA-FT was able to outperform all other methods on the malware datasets, but tested poorly on other noisy datasets (e.g. MIT-AR, Crystal Island) due to overfitting.

In realistic early classification scenarios, a single, irrevocable classification must be made, therefore it is common to report a prediction accompanied by the classifier's confidence at various times \cite{xing2008mining}. If the confidence values closely approximate the true predictive accuracy, this allows an agent to appropriately trade-off between earliness and accuracy. 
We conducted an experiment where a classifier receives higher utility for making a correct prediction with fewer observations and uses its confidences to choose the time of prediction. We note that LSTM is the state of the art and is regularly used when earliness is a factor (e.g., \cite{ma2016learning,liu2016spatio}). The results (presented in the Technical Appendix \S~C.4) show that DISC has strong performance on each domain, only comparable by LSTM. This demonstrates that DISC produces robust confidences in its predictions.

\subsection{Discussion and Limitations}
\label{sec:discussion}

We experimentally demonstrated a number of merits of our model: we achieve near-LSTM performance in most goal recognition and behaviour classification domains, as well as in an early classification task. We note that we claim no advantage over LSTMs in sequence classification and early prediction and it is not the objective of this work to demonstrate superior performance to the LSTM baseline.

One feature of our learned classifiers is that they can encode simple long-term dependencies, while n-gram classifiers cannot; in the simplest case, a unigram model does not consider the order of observations at all. DISC makes similar Markov assumptions to HMM -- that the information from any prefix of a trace can be captured by a single state -- however, DISC only considers discrete state transitions, does not model an observation emission distribution, and regularizes the size of the model. We believe these were important factors in handling noise in the data. 

A common approach to DFA-learning is to maximize training accuracy with the minimum number of DFA states. DFA-FT, which is based on this approach, excelled on the malware domains (suggesting that the DFA structure is a strong inductive bias for some real-world tasks), however, it performed poorly on many noisy datasets. The novel regularization techniques based on (non-self-loop) transitions and absorbing states introduced by DISC were crucial to learning robust DFAs which generalized to unseen data. Qualitatively, the DFAs learned by DISC were orders of magnitude smaller than those learned by DFA-FT (see Technical Appendix \S~C).

Finally, DISC assumes the traces for each label can be recognized by a DFA (or equivalently, form a regular language), which does not always hold true. In particular, DISC has limited model capacity, struggling on tasks that require large or unbounded memories, or involve counting occurrences. DISC also requires an appropriately chosen penalty on state transitions that depends on the amount of noise in the data, however, the reward for absorbing states did not require tuning in our experiments. A direction for future work is to extend DISC to handle real-valued or multi-dimensional data.

\section{Related Work}\label{sec:related_work}

We build on the large body of work concerned with learning automata from sets of traces (e.g., \cite{gold1967language,angluin1987learning,oncina1992identifying,carmel1996learning,heule2010exact,ulyantsev2015bfs,angluin2015learning,giantamidis2016learning,smetsers2018model}).
Previous approaches to learning such automata have typically constructed the prefix tree from a set of traces and employed heuristic methods or SAT solvers to minimize the resulting automaton. Here we follow a similar approach, but instead specify and realize a MILP model that is guaranteed to find optimal solutions given enough time; optimizes for a different objective function than those commonly used by previous work (see Section~\ref{sec:problem}); does not assume noise-free traces or prior knowledge of the problem (e.g., a set of DFA templates); and introduces new forms of regularization.

Some work in automata learning has also shown (some) robustness to noisy data. For instance, \citeauthor{xue2015detection} (\citeyear{xue2015detection}) combine domain-specific knowledge with domain-independent automata learning techniques and learn minimal DFAs that capture malware behaviour, with empirical results suggesting a degree of robustness to noisy data. While we eschew domain knowledge in this work, our approach allows for domain knowledge to be incorporated during the learning process. \citeauthor{ulyantsev2015bfs} (\citeyear{ulyantsev2015bfs}) also work with noisy data, but their SAT-based model assumes that at-most $k$ training instances have wrong labels, which is not a natural hyperparameter in machine learning, and does not support regularization.

Our work shares some of its motivation with previous work that has proposed to learn interpretable classifiers which favour early prediction (e.g., \cite{xing2011extracting,ghalwash2013extraction,wang2016earliness,huang2018multivariate,hsu2019multivariate}). Some work in this space appealed to discrete optimization, as we do. For instance, \citeauthor{ghalwash2013extraction} (\citeyear{ghalwash2013extraction}) leverage discrete optimization to extract salient features from time series and use those for early classification. \citeauthor{chang2012ordered} (\citeyear{chang2012ordered}) propose to learn binary classifiers that take the form of interpretable association rules; however, their approach does not consider early prediction and focuses on the binary classification task.
In our work, we specify a discrete optimization problem that yields highly structured DFA classifiers that support explanation and modification.

Relatedly, \citeauthor{bernardi2019dynamic} (\citeyear{bernardi2019dynamic}) leverage process mining techniques and perform malware detection by learning declarative models representing different families of malware. However, they only consider the binary classification task and do not consider early prediction.
Additionally, there exists a body of work that learns LTL formulae capable of discriminating between sets of traces (e.g., \cite{neider2018learning,camacho2019learning,kim2019bayesian}). These formulae can in turn be used to perform classification tasks \cite{camacho2019learning}. 
However, these works learn formulae from full traces and do not consider early prediction.

While our work was originally motivated by the goal recognition task, we have developed a general learning approach for sequence classification. % in order to address a larger collection of datasets and tasks. 
Previous work in goal and plan recognition has typically relied on rich domain knowledge (e.g., \cite{kautz1986generalized,geib2001plan,ramirez2011goal,pereira2017landmark}), thus limiting the applicability of this body of work. To leverage the existence of large datasets and machine learning techniques, some approaches to goal recognition eschew assumptions about domain knowledge and instead propose to learn models from data and use the learned models to predict an agent's goal given a sequence of observations (e.g., \cite{GeibK18,amado2018goal,polyvyanyy2020goal}). Our work partially shares its motivation with this body of work and proposes to learn models from data that offer a set of interpretability services, are optimized for early prediction, and demonstrate a capacity to generalize in noisy sequence classification settings.

There is also a body of work which applied automated planning tools to the malware detection task (e.g., \cite{sohrabi2013hypothesis,riabov2015planning}). In particular, \citeauthor{sohrabi2013hypothesis} (\citeyear{sohrabi2013hypothesis}) emphasize the importance of the robustness of a malware detection system to unreliable observations derived from network traffic, and demonstrate the robustness of their system to such observations. \citeauthor{riabov2015planning} (\citeyear{riabov2015planning}) show how robustness to noisy data can be enhanced by leveraging expert knowledge. Our learned classifiers demonstrate robustness to noisy sequence data in malware datasets and can be modified by experts to incorporate domain knowledge. \citeauthor{riabov2015planning} develop techniques which allow domain experts with no technical expertise in planning to construct models which reflect their knowledge of the domain. Such techniques could inspire methods by which domain experts can intuitively modify our learned DFA classifiers.
\section{Concluding Remarks}\label{sec:conc}

The classification of (noisy) symbolic time-series data represents a significant class of real-world problems that includes malware detection, transaction auditing, fraud detection, and a diversity of goal and behavior recognition tasks. The ability to interpret and troubleshoot these models is critical in most real-world settings. In this paper we proposed a method to address this class of problems by combining the learning of DFA sequence classifiers via MILP with Bayesian inference. Our approach introduced novel automata-learning techniques crucial to addressing regularization, efficiency, and early classification. Critically, the resulting DFA classifiers offer a set of interpretability services that include explanation, counterfactual reasoning, verification of properties, and human modification. Our implemented system, DISC, achieves similar performance to LSTMs and superior performance to HMMs and n-grams on a set of synthetic and real-world datasets, with the important advantage of being interpretable.

\subsection*{Acknowledgements}
We gratefully acknowledge funding from the Natural Sciences and Engineering Research Council of Canada (NSERC), the Canada CIFAR AI Chairs Program, and Microsoft Research. Rodrigo also gratefully acknowledges funding from ANID (Becas Chile). Resources used in preparing this research were provided, in part, by the Province of Ontario, the Government of Canada through CIFAR, and companies sponsoring the Vector Institute for Artificial Intelligence \url{www.vectorinstitute.ai/partners}.

%%\begin{small}
%%{\fs{9}
%{\small

\bibliography{main}
%}
%\end{small}

\onecolumn
\newpage
\appendix

% Title
{
\begin{center}
  \Huge{Technical Appendix}
  \vspace{2ex}
\end{center}
}

\addcontentsline{toc}{section}{} % Add the appendix text to the document TOC
\part{} % Start the appendix part
\parttoc % Insert the appendix TOC

In Appendix~A, we provide further details concering our procedure to learn a DFA-based classifier from a training set. In Appendix~B, we outline a simple natural language generation approach for counterfactual explanation, present samples of learned DFA classifiers from our experiments, and provide exposition of Linear Temporal Logic. In Appendix~C, we provide additional details of our experimental setup and our datasets and present additional experimental results (including results from early and multi-label classification experiments).

\sectionfont{\raggedright}

\section{Learning DFAs from Training Data}
\label{sup:learn_dfa}

In this appendix, we provide further details concerning our procedure to learn a DFA-based classifier from a training set $\mathcal{T} = \{(\tau_1, c_1), \ldots, (\tau_N, c_N)\}$. Recall that, for each possible label $c \in \classLabels$, we train a separate DFA, $\mathcal{M}_c$, responsible for recognizing traces with label $c$. We then use those DFAs to compute a probability distribution for online classification of partial traces.

\subsection{From Training Data to Prefix Trees}

The first step to learning a DFA is to construct a Prefix Tree (PT). Algorithm~\ref{fig:algorithm_PT} shows the pseudo-code to do so. It receives the training set $\mathcal{T}$ and the label $c^+$ of the positive class. It returns the PT for that training set and class label. It also labels each PT node with the costs associated with classifying that node as positive and negative, respectively. 
That cost depends on the length of the trace and its label, and it is computed using the function \texttt{add\_cost}().

\begin{algorithm}
\caption{Converting Training Data into Prefix Trees}
\label{fig:algorithm_PT}
\begin{algorithmic}[1]%
\Function{get\_prefix\_tree}{$\mathcal{T}$, $c^+$}
    \State $r \gets $ create\_root\_node()
    \For{$(\tau,c) \in \mathcal{T}$} 
        \State $n \gets r$
        \State add\_cost($n$, $c=c^+$, $|\tau|$)
        \For{$\sigma \in \tau$}
            \If {\textbf{not} has\_child($n$,$\sigma$)}
                \State add\_child($n$,$\sigma$)
            \EndIf
            \State $n \gets $ get\_child($n$,$\sigma$)
            \State add\_cost($n$, $c=c^+$, $|\tau|$)
        \EndFor
    \EndFor
    \State \Return $r$
\EndFunction\end{algorithmic}%
\end{algorithm}

\subsection{From Prefix Trees to DFAs}
\label{supp:prefix_tree_to_dfa}

% Model overview
We now discuss the MILP model we use to learn a DFA given a PT. The complete MILP model follows.
{%
\begingroup%
\allowdisplaybreaks
%\small
\begin{align}%
\min\; & \sum_{n \in N} c_n + \lambda_e \sum_{q \in Q}\sum_{\sigma \in \Sigma} e_{q,\sigma} + \lambda_t \sum_{n \in N} t_n \tag{\texttt{MILP}} \label{mip:problem}\\
s.t.\; 
& \sum_{q \in Q} x_{n,q} = 1 & \forall n \in N  \label{mip:one_state}\\
& x_{r,0} = 1 & \label{mip:q0} \\
& \sum_{q' \in Q} \delta_{q,\sigma,q'} = 1 & \forall q \in Q, \sigma \in \Sigma   \label{mip:one_transition}\\
& \delta_{q,\sigma,q} = 1 & \forall q \in T, \sigma \in \Sigma   \label{mip:terminal_nodes}\\
& x_{p(n),q} + x_{n,q'} -1 \leq \delta_{q,s(n),q'}  & \forall n \in N \setminus \{r\}, q \in Q, q' \in Q \label{mip:feasible_assignment}\\
& c_n = \lambda^+ \sum_{q\in F} c^+(n) x_{n,q} + \lambda^- \sum_{q\in Q \setminus F} c^-(n) x_{n,q} & \forall n \in N \label{mip:cost_pred}\\
& e_{q,\sigma} = \sum_{q' \in Q\setminus\{q\}} \delta_{q,\sigma,q'} & \forall q \in Q, \sigma \in \Sigma \label{mip:cost_edge}\\
& t_n = \sum_{q \in Q \setminus T} x_{n,q} & \forall n \in N \label{mip:cost_non_terminal}\\
& x_{n,q} \in \{0,1\} &\forall n \in N, q \in Q \label{mip:dom_x}\\
& \delta_{q,\sigma,q'} \in \{0,1\} & \forall q \in Q, \sigma \in \Sigma, q' \in Q \label{mip:dom_delta}\\
& c_n \in \reals & \forall n \in N  \label{mip:dom_c}\\
& e_{q,\sigma} \in \reals & \forall q \in Q, \sigma \in \Sigma  \label{mip:dom_e}\\
& t_n \in \reals & \forall n \in N \label{mip:dom_t}
\end{align}%
\endgroup%
}%%

This model learns a DFA over a vocabulary $\Sigma$ with at most $q_{\max}$ states. From those potential states $Q$, we set state $0$ to be the initial state $q_0$ and predefine a set of accepting states $F\subset Q$ and a set of terminal states $T\subset Q$. We also use the following notation to refer to nodes in the PT: $r$ is the root node, $p(n)$ is the parent of node $n$, $s(n)$ is the symbol that caused node $p(n)$ to transition to node $n$, $c^+(n)$ is the cost associated with predicting node $n$ as positive, $c^-(n)$ is the cost associated with predicting node $n$ as negative, and $N$ is the set of all PT nodes. The model also has hyperparameters $\lambda_e$ and $\lambda_t$ to weight our regularizers and hyperparameters $\lambda^+$ and $\lambda^-$ to penalize misclassifications of positive and negative examples differently (in the case where the training data is imbalanced).

% Variables
The idea behind our model is to assign DFA states to each node in the tree. Then, we look for an assignment that is feasible (i.e., it can be produced by a deterministic DFA) and optimizes a particular objective function---which we describe later. The main decision variables are $x_{n,q}$ and $\delta_{q,\sigma,q'}$, both binary. Variable $x_{n,q}$ is $1$ %one 
iff node $n \in N$ is assigned the DFA state $q \in Q$. Variable $\delta_{q,\sigma,q'}$ is $1$ %one 
iff the DFA transitions from state $q\in Q$ to state $q'\in Q$ given symbol $\sigma \in \Sigma$. Note that $c_n$, $e_{q,\sigma}$, and $t_n$ are auxiliary (continuous) variables used to compute the cost of the DFAs.

% Constraints
Constraint \eqref{mip:one_state} ensures that only one DFA state is assigned to every PT node and constraint \eqref{mip:q0} forces the root node to be assigned to $q_0$. Constraint \eqref{mip:one_transition} ensures that the DFA is deterministic and constraint \eqref{mip:terminal_nodes} makes the terminal nodes sink nodes. Finally, constraint \eqref{mip:feasible_assignment} ensures that the assignment can be emulated by the DFA. The rest of the constraints compute the cost of solutions and the domain of the variables. In particular, note that the objective function minimizes the prediction error using $c_n$, the number of transitions between different DFA states using $e_{q,\sigma}$, and the occupancy of non-terminal states using $t_n$.

This model has $O(|N||Q| + |\Sigma||Q|^2)$ decision variables and $O(|N||Q|^2 + |\Sigma||Q|)$ constraints.

\subsection{Derivation of the Posterior Probability Distribution over the Set of Class Labels}

Recall the following assumptions: {(1) the classification decisions $\mathrm{D}_c$ for $c \in \classLabels$ are conditionally independent, given the true label $c^{*}$} and (2) $p(D_c | c^*)$ only depends on whether $c = c^*$.

For each $c'$, we compute the posterior probability of $c^* = c'$ to be
\begin{align*}
    &\hspace{5mm} p(c^* = c' | \{ \mathrm{D}_c : c \in \classLabels \} ) \\
    &\propto p(c^* = c') * p(\{ \mathrm{D}_c : c \in \classLabels \} | c^* = c') \hspace{5mm} \text{(using Bayes' rule)} \\
    &= p(c^* = c') * \prod_{c \in \classLabels} {p(D_c | c^* = c')} \hspace{5mm} \text{(using (1))} \\ 
    &= p(c^* = c') * p(D_{c'} | c^* = c') * \prod_{c \in \classLabels \setminus \{c'\}} {p(D_c | c^* \neq c)} \hspace{5mm} \text{(using (2))} \\  
    &\propto \frac{p(c^* = c') * p(D_{c'} | c^* = c')} { p(D_{c'} | c^* \neq c') } \hspace{5mm} \text{(dividing through by the constant} \prod_{c \in \classLabels} {p(D_c | c^* \neq c)}\text{)}
\end{align*}

\section{Interpretability}
\label{supp:inter}

\subsection{Natural Language Generation for Counterfactual Explanation}\label{supp:NLG_counter}

In Section~\ref{sec:interp} of our paper, we discussed counterfactual explanations, which are useful in cases where a classifier does not return a positive classification for a trace.
Here we describe a simple algorithm that transforms a counterfactual explanation (comprising a sequence of edit operations - see Definition~\ref{def:counterfactual_exp} for details) to an English sentence.
We define three edit operations over strings: \textsc{Replace}($s$, $c_1$, $c_2$) replaces the first occurrence of the character $c_1$ in the string $s$ with the character $c_2$; \textsc{Insert}($s$, $c_1$, $c_2$) inserts the character $c_1$ after the first occurrence of the character $c_2$ in the string $s$; \textsc{Delete}($s$, $c_1$) removes the first occurrence of the character $c_1$ from the string $s$.

Algorithm~\ref{alg:NLG_counterfactual} accepts as input a sequence of edit operations $e_1, e_2, \ldots, e_n$ (where $e_i$ is either a replace, insert, or delete operation), and returns a string representing an English sentence encoding the counterfactual explanation for some trace $\tau$. Redundant ``and"s are removed from the resulting string. We do not consider multiple occurrences of the same character in a single string but this can be easily handled. $e_i.args[i]$ is assumed to return the $i + 1$th argument of an edit operation $e_i$ and $e_i.type$ is assumed to return the type of the edit operation (e.g., \textsc{Replace}). \textsc{Concatenate}($s_1$, $s_2$) appends the string $s_2$ to the suffix of the string $s_1$. We further assume that connectives (e.g., `and') are added between the substrings representing the edit operations.

\begin{algorithm}[h!]
\begin{small}
\floatname{algorithm}{Algorithm}
\caption{Natural Language Generation for Counterfactual Explanation}
\label{alg:NLG_counterfactual}
\begin{algorithmic}[1]

\Require A sequence of edit operations $E = e_1, e_2, \ldots, e_n$
% and a trace $\tau$

\State  $s \gets $ \textit{``The binary classifier would have accepted the trace"}

\State For $e$ in E:

\State \quad If $e$.type == \textsc{Replace}
\State \quad\quad \textsc{Concatenate}($s$, \textit{`` had $e.args[2]$ been observed instead of $e.args[1]$"})
    
\State \quad If $e_i$.type == \textsc{Insert}
\State \quad\quad \textsc{Concatenate}($s$, \textit{`` had $e.args[2]$ been observed following the observation of $e.args[1]$"})

\State \quad If $e$.type == \textsc{Delete}
\State \quad\quad \textsc{Concatenate}($s$, \textit{`` had $e.args[1]$ been removed from the trace"})

\State RETURN $s$

\end{algorithmic}
\label{alg1}
\end{small}
\end{algorithm}

For example, using the DFA depicted in Figure~\ref{fig:running_example}, if $\tau = $ (\textbf{A, H2, H1,} {\faMale}) then a possible counterfactual explanation is the edit operation \textsc{Replace}($\tau$, \faMale, \coffee) which transforms (\textbf{A, H2, H1,} {\faMale}) to (\textbf{A, H2, H1,} {\coffee}). Given the edit operation \textsc{Replace}($\tau$, \faMale, \coffee), Algorithm 2 returns the string \textit{``The binary classifier would have accepted the trace had \coffee been observed instead of \faMale"}. 

\subsection{Samples of Learned DFA Classifiers}\label{supp:DFA_examples}

In this appendix we present a number of examples of DFAs learned by DISC from our experimental evaluation in Section~\ref{sec:exp} of our paper. As discussed in Section~\ref{sec:interp}, our purpose in this work is to highlight the breadth of interpretability services afforded by DFA classifiers via their relationship to formal language theory. The effectiveness of a particular interpretability service is user-, domain-, and even task-specific and is best evaluated in the context of individual domains. Moreover, the DFAs presented in this appendix require familiarity with the domain in question and are therefore best suited for domain experts.

\subsubsection{Malware}

We present two DFAs learned via DISC from the real-world \textit{malware} datasets. Figure~\ref{fig:supp_example_DFA_malware_10_state} depicts a DFA classifier for BatteryLow that detects whether a trace of Android system calls was issued by the malware family \textit{DroidKungFu4}. The maximum number of states is limited to 10. The trace $\tau = $ (\textit{sendto, epoll\_wait, recvfrom, gettid, getpid, read}) is rejected by the depicted \textit{DroidKungFu4} DFA classifier. This can be seen by starting at the initial state of the DFA, $q_0$, and mentally following the DFA transitions corresponding to the symbols in the trace.  The exercise of following the symbols of the trace transition through the DFA can be done by anyone. For the domain expert, the symbols have meaning (and can be replaced by natural language words that are even more evocative, as necessary). In this trace we see that rather than stopping at accepting state $q_6$, the trace transitions in the DFA to $q_5$, a non-accepting state. One %possible 
counterfactual explanation that our system generates
to address what changes could result in a positive classification is: ``The binary classifier would have accepted the trace had \textit{read} been removed from the trace" (per the algorithm in Appendix~\ref{supp:NLG_counter}).

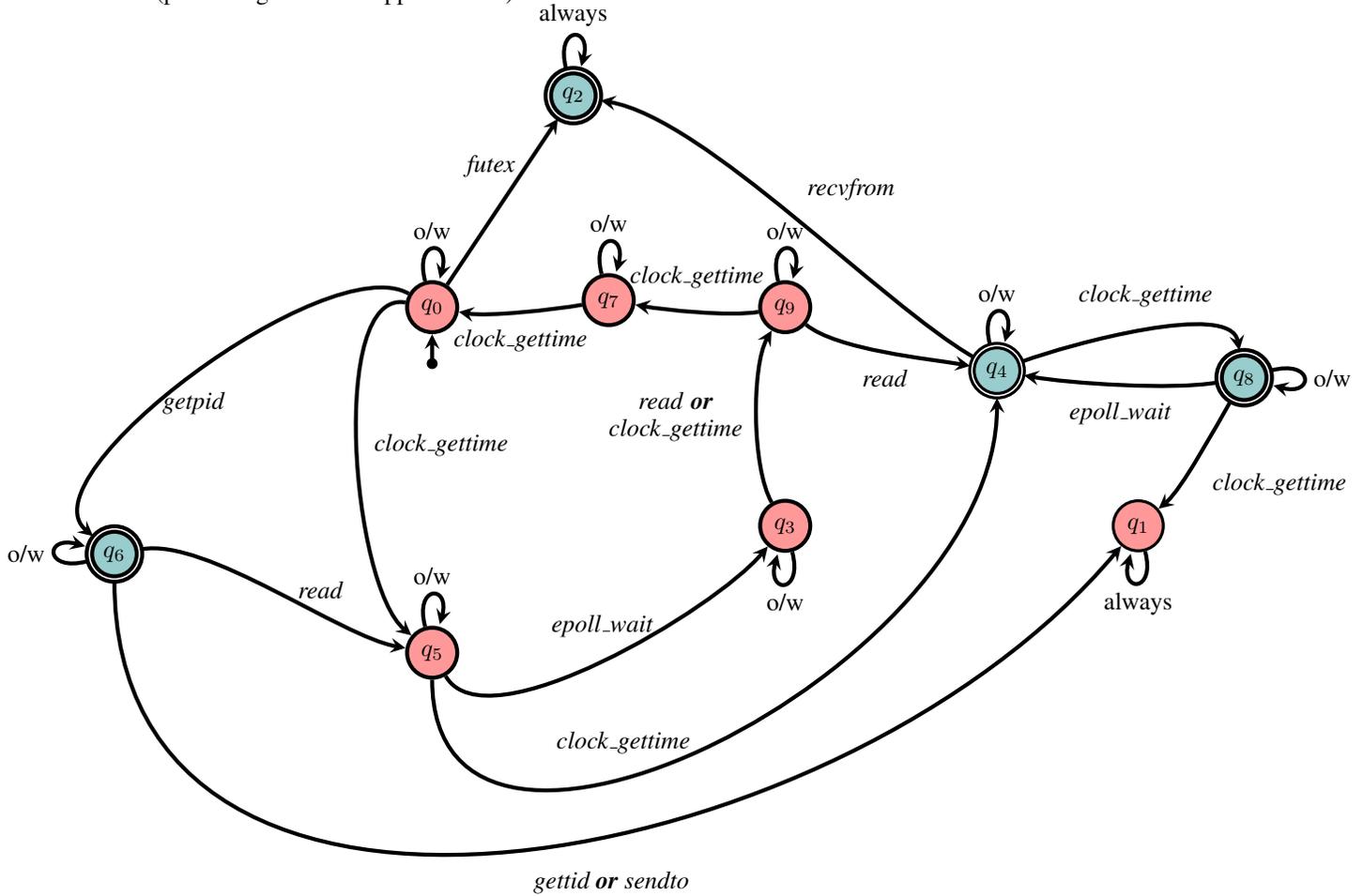
\begin{figure}[h!]
    \centering
    \vspace{-6mm}
    \begin{tikzpicture}[node distance=2cm,on grid,every initial by arrow/.style={ultra thick,->, >=stealth}, initial text={}]%
  \node[ultra thick,state,fill=red!40!white,minimum size=.5cm] (q_0) at (-5/2,2) {$q_0$};
  \node[ultra thick,state,fill=red!40!white,minimum size=.5cm]  (q_3) at (5/2,-1.1) {$q_3$};
  
  \node[very thick,state, fill=red!40!white,minimum size=.5cm]  (q_1) at (15/2,-1.1) {$q_1$};
  
  \node[ultra thick,state,accepting,fill=teal!40!white,minimum size=.5cm]  (q_8) at (18/2,1) {$q_8$};
  \node[ultra thick,state,accepting,fill=teal!40!white,minimum size=.5cm]  (q_2) at (-1/2,5) {$q_2$};
  %\node[very thick,state,accepting, fill=teal!40!white,minimum size=.5cm]  (q_4) at (15/2,-1.4) {$q_4$};
  \node[very thick,state,accepting, fill=teal!40!white,minimum size=.5cm]  (q_4) at (11/2,1.1) {$q_4$};
  \node[ultra thick,state,fill=red!40!white,minimum size=.5cm]  (q_5) at (-5/2,-2.9) {$q_5$};
  
  \node[ultra thick,state,fill=red!40!white,minimum size=.5cm]  (q_7) at (0,2.1) {$q_7$};
  
  \node[ultra thick,state,fill=red!40!white,minimum size=.5cm]  (q_9) at (5/2,2) {$q_9$};
  
  \node[ultra thick,state,accepting,fill=teal!40!white,minimum size=.5cm]  (q_6) at (-14/2,-1.5) {$q_6$};
  
%   \node[ultra thick,state,fill=red!40!white,minimum size=.5cm] (q_0) at (-9/2,0) {$q_6$};

  % Initial node
  \path[ultra thick,->, >=stealth] (-5/2,1.2) edge node [right] {} (q_0);
  \draw[fill=black] (-5/2,1.2) circle (0.07);
  
  % Loops
  \path[->] (q_0) edge [ultra thick,->, >=stealth,loop above] node { \text{o/w}} ();
  \path[->] (q_1) edge [ultra thick,->, >=stealth,loop below] node { \text{always}} ();
  \path[->] (q_2) edge [ultra thick,->, >=stealth,loop above] node { \text{always}} ();
  \path[->] (q_3) edge [ultra thick,->, >=stealth,loop below] node { \text{o/w}} ();
  \path[->] (q_4) edge [ultra thick,->, >=stealth,loop above] node { \text{o/w}} ();
  \path[->] (q_5) edge [ultra thick,->, >=stealth,loop above] node { \text{o/w}} ();
  \path[->] (q_6) edge [ultra thick,->, >=stealth,loop left] node { \text{o/w}} ();
  \path[->] (q_7) edge [ultra thick,->, >=stealth,loop above] node { \text{o/w}} ();
  \path[->] (q_8) edge [ultra thick,->, >=stealth,loop right] node { \text{o/w}} ();
  \path[->] (q_9) edge [ultra thick,->, >=stealth,loop above] node { \text{o/w}} ();
  
  % Edges
  
  \path[ultra thick,->, >=stealth,text width=1cm] (q_0) edge node [above=0.2] { \textit{futex} } (q_2);
  
%   \draw[ultra thick,->, >=stealth, out=10, in=170, looseness=0.6] (q_0) to node [right=0.2] {\text{{\textit{epoll\_wait}}}} (q_3);
  
  \draw[ultra thick,->, >=stealth, out=10, in=170, looseness=0.6] (q_6) to node [right=0.2] {\text{{\textit{read}}}} (q_5);
  
  \draw[ultra thick,->, >=stealth, out=-40, in=170, looseness=0.6] (q_9) to node [below=0.1] {\text{{\textit{read}}}} (q_4);
  
  \draw[ultra thick,->, >=stealth, out=270, in=230, looseness=1.1] (q_6) to node [below right=0.2] {\text{{\textit{gettid \textbf{or} sendto}}}} (q_1);
  
  \draw[ultra thick,->, >=stealth, out=270, in=270, looseness=1.1] (q_5) to node [left=0.2] {\text{{\textit{clock\_gettime}}}} (q_4);
  
%   \draw[ultra thick,->, >=stealth, out=120, in=350, looseness=0.6] (q_2) to node [below=0.07] {\text{{\textit{writev}}}} (q_1);
  
%   \draw[ultra thick,->, >=stealth, out=190, in=-10, looseness=0.6] (q_3) edge node [below=0.1] {\text{{\textit{read}}}} (q_0);
  
%   \draw[ultra thick,->, >=stealth, out=350, in=30, looseness=0.6] (q_4) edge node [below=0.1] {\text{{\textit{read}}}} (q_1);
  
%   \path[ultra thick,->, >=stealth,text width=1cm] (q_3) edge node [above=0.2] { \textit{recvfrom} } (q_1);

%   \draw[ultra thick,->, >=stealth, out=100, in=-180, looseness=0.6] (q_3) edge node [below=0.1] {\text{{\textit{read}}}} (q_4);
  
  \draw[ultra thick,->, >=stealth, out=170, in=-220, looseness=0.6] (q_0) edge node [right=0.1] {\text{{\textit{clock\_gettime}}}} (q_5);
  
  \draw[ultra thick,->, >=stealth, out=150, in=-220, looseness=0.6] (q_0) edge node [below=0.1] {\text{{\textit{getpid}}}} (q_6);
  
%   \draw[ultra thick,->, >=stealth, out=120, in=-180, looseness=0.6] (q_5) edge node [below=0.1] {\text{{\textit{read}}}} (q_2);
  
  \path[ultra thick,->, >=stealth, out=120, in=-120, looseness=0.6] (q_3) edge node [left=0, text width=2cm, align=center] {{\textit{read \textbf{or} \\ clock\_gettime}}} (q_9);
  
%   \newline\vspace{1em}
  
  \draw[ultra thick,->, >=stealth, out=300, in=-130, looseness=0.6] (q_5) edge node [above=0.2] {\text{{\textit{epoll\_wait}}}} (q_3);
  
%   \draw[ultra thick,->, >=stealth, out=190, in=-10, looseness=0.6] (q_3) edge node [below=0.1] {\text{{\textit{read}}}} (q_2);
  
  \draw[ultra thick,->, >=stealth, out=190, in=-10, looseness=0.6] (q_7) edge node [below=0.1] {\text{{\textit{clock\_gettime}}}} (q_0);
  
  \draw[ultra thick,->, >=stealth, out=190, in=-10, looseness=0.6] (q_9) edge node [above=0.2] {\text{{\textit{clock\_gettime}}}} (q_7);
  
  \draw[ultra thick,->, >=stealth, out=150, in=-10, looseness=0.6] (q_4) edge node [above right=0.1] {\text{{\textit{recvfrom}}}} (q_2);
  
  \draw[ultra thick,->, >=stealth, out=190, in=-10, looseness=0.6] (q_8) edge node [below=0.1] {\text{{\textit{epoll\_wait}}}} (q_4);
  
  \draw[ultra thick,->, >=stealth, out=240, in=40, looseness=0.6] (q_8) edge node [below right=0.1] {\text{{\textit{clock\_gettime}}}} (q_1);
  
  \draw[ultra thick,->, >=stealth, out=20, in=100, looseness=0.6] (q_4) to node [above=0.2] {\text{{\textit{clock\_gettime}}}} (q_8);
  
%   \path[ultra thick,->, >=stealth] (q_3) edge node [right=0.2] {\textbf{\small A or B}} (q_1);
  
%   \path[ultra thick,->, >=stealth, out=2, in=190, looseness=0.8] (q_3) edge node [below] {\textit{{clock\_gettime}}} (q_2);
  
%   \path[ultra thick,->, >=stealth, out=90, in=-80, looseness=0.8] (q_2) edge node [above right] {\textit{getuid32}} (q_4);

\end{tikzpicture}%

% delta(0,21) = 2
% futex
% delta(0,0) = 5
%clock_gettime
% delta(0,4) = 6
%getpid
% delta(3,0) = 9
%clock_gettime
% delta(3,2) = 9
%read
% delta(4,6) = 2
%recvfrom
% delta(4,2) = 6
%read
% delta(4,0) = 8
%clock_gettime
% delta(5,1) = 3
%epoll\_wait
% delta(5,0) = 4
%clock_gettime
% delta(6,16) = 1
%gettid
% delta(6,8) = 1
%sendto
% delta(6,2) = 5
%read
% delta(7,0) = 0
%clock_gettime
% delta(8,0) = 1
%clock\_gettime
% delta(8,1) = 4
%epoll\_wait
% delta(9,2) = 4
%read
% delta(9,0) = 7
%clock\_gettime
% q_0 = 0
    \caption{A DFA learned in our experiments from the BatteryLow dataset by limiting the maximum number of states to 10. A decision is provided after each new observation based on the current state: yes for the blue accepting state, and no for the red, non-accepting states. “o/w” (otherwise) stands for all symbols that do not appear on outgoing edges from a state. “always” stands for all symbols.}
    \label{fig:supp_example_DFA_malware_10_state}
\end{figure}

For comparison, Figure~\ref{fig:supp_example_DFA_malware_5_state} presents a smaller DFA for BootCompleted, learned with the maximum number of states limited to 5. (This DFA was not used in our experiments.) The DFA detects whether a trace was issued by the malware family \textit{DroidDream}. The trace (here truncated, as subsequent observations do not affect the classification decision) $\tau = $ (\textit{clock\_gettime, epoll\_wait, clock\_gettime, clock\_gettime, getpid, writev, ...}) is rejected by the DFA. One counterfactual explanation that our system generates
to address what changes could result in a positive classification is: ``The binary classifier would have accepted the trace had \textit{getuid32} been observed instead of \textit{writev}".

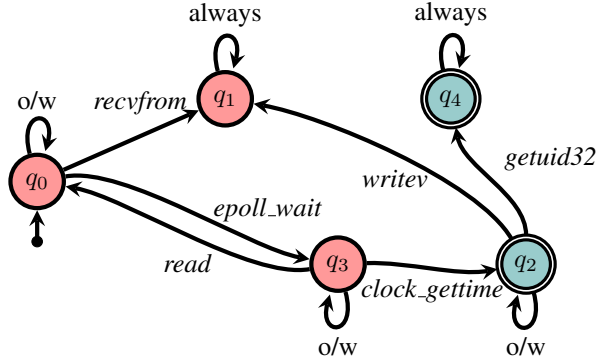
\begin{figure}[h!]
    \centering
    %\vspace{-6mm}
    \begin{tikzpicture}[node distance=2cm,on grid,every initial by arrow/.style={ultra thick,->, >=stealth}, initial text={}]%
  \node[ultra thick,state,fill=red!40!white,minimum size=.5cm] (q_0) at (-3/2,0) {$q_0$};
  \node[ultra thick,state,fill=red!40!white,minimum size=.5cm]  (q_3) at (5/2,-1.1) {$q_3$};
  \node[very thick,state,accepting, fill=teal!40!white,minimum size=.5cm]  (q_2) at (10/2,-1.1) {$q_2$};
  \node[ultra thick,state,fill=red!40!white,minimum size=.5cm]  (q_1) at (2/2,1.1) {$q_1$};
  %\node[very thick,state,accepting, fill=teal!40!white,minimum size=.5cm]  (q_4) at (15/2,-1.4) {$q_4$};
  \node[very thick,state,accepting, fill=teal!40!white,minimum size=.5cm]  (q_4) at (8/2,1.1) {$q_4$};

  % Initial node
  \path[ultra thick,->, >=stealth] (-3/2,-0.8) edge node [right] {} (q_0);
  \draw[fill=black] (-3/2,-0.8) circle (0.07);
  
  % Loops
  \path[->] (q_0) edge [ultra thick,->, >=stealth,loop above] node { \text{o/w}} ();
  \path[->] (q_1) edge [ultra thick,->, >=stealth,loop above] node { \text{always}} ();
  \path[->] (q_2) edge [ultra thick,->, >=stealth,loop below] node { \text{o/w}} ();
  \path[->] (q_3) edge [ultra thick,->, >=stealth,loop below] node { \text{o/w}} ();
  \path[->] (q_4) edge [ultra thick,->, >=stealth,loop above] node { \text{always}} ();
  
  % Edges

 % from 0 to 1
%   'writev"/>
% 'gettid"/>
% 'stat64"/>
% 'futex"/>
% 'recvfrom"/>
% 'sendto"/>'
  
%   \path[ultra thick,->, >=stealth] (q_0) edge node [below left] {\text{..}} (q_2);

 % from 0 to 3
 %'epoll_wait"/>'
 %getpid"/>'
  
  % from 3 to 0
  %epoll\_wait
  % read
  
  % from 3 to 4
  
  % from 4 to 1
  
  % from 4 to 2
  
%   \\ \textit{writev} \\ \textit{gettid} \\ \textit{stat64} \\ \textit{futex}
  
  \path[ultra thick,->, >=stealth,text width=1cm] (q_0) edge node [above=0.2] { \textit{recvfrom} } (q_1);
  
  \draw[ultra thick,->, >=stealth, out=10, in=170, looseness=0.6] (q_0) to node [right=0.2] {\text{{\textit{epoll\_wait}}}} (q_3);
  
  \draw[ultra thick,->, >=stealth, out=120, in=350, looseness=0.6] (q_2) to node [below=0.07] {\text{{\textit{writev}}}} (q_1);
  
  \draw[ultra thick,->, >=stealth, out=190, in=-10, looseness=0.6] (q_3) edge node [below=0.1] {\text{{\textit{read}}}} (q_0);
  
%   \path[ultra thick,->, >=stealth] (q_3) edge node [right=0.2] {\textbf{\small A or B}} (q_1);
  
  \path[ultra thick,->, >=stealth, out=2, in=190, looseness=0.8] (q_3) edge node [below] {\textit{{clock\_gettime}}} (q_2);
  
  \path[ultra thick,->, >=stealth, out=90, in=-80, looseness=0.8] (q_2) edge node [above right] {\textit{getuid32}} (q_4);

\end{tikzpicture}%

% 0,1,2,0,0,0,6
% 0,1,0,0,0,0,3,0,4,5
% clock_gettime, epoll\_wait, clock_gettime, clock_gettime,clock_gettime,clock_gettime, getpid, getuid32

% delta(0,15) = 1 writev
% delta(0,16) = 1 gettid
% delta(0,17) = 1 stat64
% delta(0,21) = 1 futex
% delta(0,6) = 1 recvfrom
% delta(0,8) = 1 sendto

% delta(0,1) = 3 epoll\_wait
% delta(0,4) = 3 getpid

% delta(3,1) = 0 epoll\_wait
% delta(3,2) = 0 read

% delta(3,0) = 2 clock_gettime
% delta(3,6) = 2 recvfrom

% delta(2,10) = 1 mmap2
% delta(2,15) = 1  writev
% delta(2,16) = 1 gettid
% delta(2,18) = 1 mprotect

% delta(2,5) = 4 getuid32
    \caption{A DFA learned from the BootCompleted dataset by limiting the maximum number of states to 5. A decision is provided after each new observation based on the current state: yes for the blue accepting state, and no for the red, non-accepting states. “o/w” (otherwise) stands for all symbols that do not appear on outgoing edges from a state. “always” stands for all symbols.}
    \label{fig:supp_example_DFA_malware_5_state}
\end{figure}

Note that while the 5-state DFA may be more interpretable to humans than the 10-state DFA, the 10-state DFA can model more complex patterns in the data. Indeed, during the course of our experiments with the malware datasets, we found that setting $q_{\max} = 10$ achieved superior performance to $q_{\max} = 5$. 

\subsubsection{Crystal Island} Figure~\ref{fig:supp_example_DFA_crystal_10_state} depicts a DFA classifier learned from the Crystal Island dataset that detects whether a trace of player actions was performed in order to achieve the goal \textit{Talked-to-Ford}. The maximum number of states is limited to 5. 

Consider the trace $\tau = $ (\textit{pickup banana, move outdoors (2a), move outdoors (2b), open door infirmary bathroom, move outdoors (3a), move hall}), which is rejected by the depicted \textit{Talked-to-Ford} DFA classifier. One possible 
counterfactual explanation to result in a positive trace is: ``The binary classifier would have accepted the trace had \textit{move sittingarea} been observed following the observation of \textit{move hall}". Additionally, a \emph{necessary condition} for this DFA to accept is that either \textit{talk ford} or \textit{move sittingarea} is observed --- or equivalently, the LTL property $\ltleventually{}$ (\textit{talk ford} $\lor$ \textit{move sittingarea}). This LTL formula is entailed by the DFA.

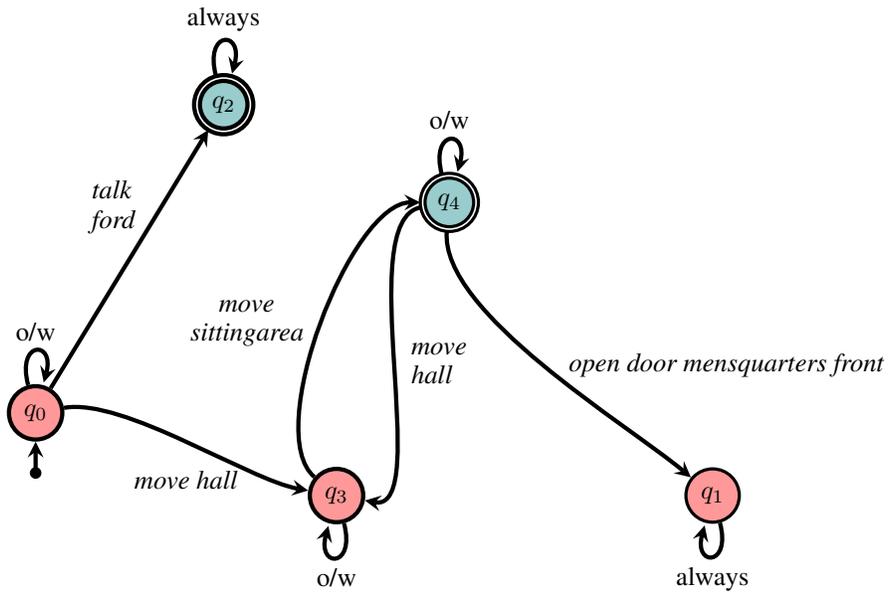
\begin{figure}[h!]
    \centering
    \vspace{-6mm}
    \begin{tikzpicture}[node distance=2cm,on grid,every initial by arrow/.style={ultra thick,->, >=stealth}, initial text={}]%
  \node[ultra thick,state,fill=red!40!white,minimum size=.5cm] (q_0) at (-3/2,0) {$q_0$};
  \node[ultra thick,state,fill=red!40!white,minimum size=.5cm]  (q_3) at (5/2,-1.1) {$q_3$};
  \node[very thick,state, fill=red!40!white,minimum size=.5cm]  (q_1) at (15/2,-1.1) {$q_1$};
  \node[ultra thick,state,accepting,fill=teal!40!white,minimum size=.5cm]  (q_2) at (2/2,4.1) {$q_2$};
  %\node[very thick,state,accepting, fill=teal!40!white,minimum size=.5cm]  (q_4) at (15/2,-1.4) {$q_4$};
  \node[very thick,state,accepting, fill=teal!40!white,minimum size=.5cm]  (q_4) at (8/2,2.8) {$q_4$};

  % Initial node
  \path[ultra thick,->, >=stealth] (-3/2,-0.8) edge node [right] {} (q_0);
  \draw[fill=black] (-3/2,-0.8) circle (0.07);
  
  % Loops
  \path[->] (q_0) edge [ultra thick,->, >=stealth,loop above] node { \text{o/w}} ();
  \path[->] (q_1) edge [ultra thick,->, >=stealth,loop below] node { \text{always}} ();
  \path[->] (q_2) edge [ultra thick,->, >=stealth,loop above] node { \text{always}} ();
  \path[->] (q_3) edge [ultra thick,->, >=stealth,loop below] node { \text{o/w}} ();
  \path[->] (q_4) edge [ultra thick,->, >=stealth,loop above] node { \text{o/w}} ();
  
  % Edges
  
  \path[ultra thick,->, >=stealth,text width=1cm] (q_0) edge node [above=0.2] { \textit{talk ford} } (q_2);
  
  cur-loc-livingquarters-sittingarea
% cur-action-movetoloc-livingquarters-robertsroom
  
  \draw[ultra thick,->, >=stealth, out=10, in=170, looseness=0.6] (q_0) to node [below=0.25] {\text{{\textit{move hall}}}} (q_3);
  
%   \draw[ultra thick,->, >=stealth, out=120, in=350, looseness=0.6] (q_2) to node [below=0.07] {\text{{\textit{writev}}}} (q_1);
  
%   \draw[ultra thick,->, >=stealth, out=190, in=-10, looseness=0.6] (q_3) edge node [below=0.1] {\text{{\textit{read}}}} (q_0);
  
  \draw[ultra thick,->, >=stealth, out=265, in=-220, looseness=0.6] (q_4) edge node [right=0.2] {\text{{\textit{open door mensquarters front}}}} (q_1);
  
  \draw[ultra thick,->, >=stealth, out=190, in=-10, looseness=0.6] (q_4) edge node [right=0.1, text width=1cm] {{\textit{move \\ hall}}} (q_3);
  
  \draw[ultra thick,->, >=stealth, out=140, in=-180, looseness=0.6] (q_3) edge node [left=0.1, text width=1.5cm, align=center] {{\textit{move \\ sittingarea}}} (q_4);
  
%   \path[ultra thick,->, >=stealth] (q_3) edge node [right=0.2] {\textbf{\small A or B}} (q_1);
  
%   \path[ultra thick,->, >=stealth, out=2, in=190, looseness=0.8] (q_3) edge node [below] {\textit{{clock\_gettime}}} (q_2);
  
%   \path[ultra thick,->, >=stealth, out=90, in=-80, looseness=0.8] (q_2) edge node [above right] {\textit{getuid32}} (q_4);

\end{tikzpicture}%
    \caption{A DFA learned from the Crystal Island dataset, limiting the maximum number of states to 5. A decision is provided after each new observation based on the current state: yes for the blue accepting state, and no for the red, non-accepting states. “o/w” (otherwise) stands for all symbols that do not appear on outgoing edges from a state. “always” stands for all symbols.}
    \label{fig:supp_example_DFA_crystal_10_state}
\end{figure}

\subsubsection{StarCraft} Figure~\ref{fig:supp_example_DFA_starcraft_10_state} depicts a DFA classifier learned from the StarCraft dataset that detects whether a trace of actions was generated by the StarCraft-playing agent \textit{EconomyMilitaryRush}. The maximum number of states is limited to 10. 

The trace $\tau = $ (\textit{move produce, produce, move produce, move, move, move, harvest, harvest, harvest, move produce, move produce, attack move move }) is rejected by the depicted \textit{EconomyMilitaryRush} DFA classifier. A %possible 
counterfactual explanation to result in a positive trace is: ``The binary classifier would have accepted the trace had \textit{harvest move} been observed instead of \textit{attack move move}". A \emph{necessary 
condition} for the DFA to accept is that either \textit{move produce} or \textit{harvest produce} is observed in the trace. Furthermore, every trace starting with \textit{harvest produce} will be accepted. As discussed in Section~\ref{sec:interp}, these properties can be automatically extracted from the DFAs. 

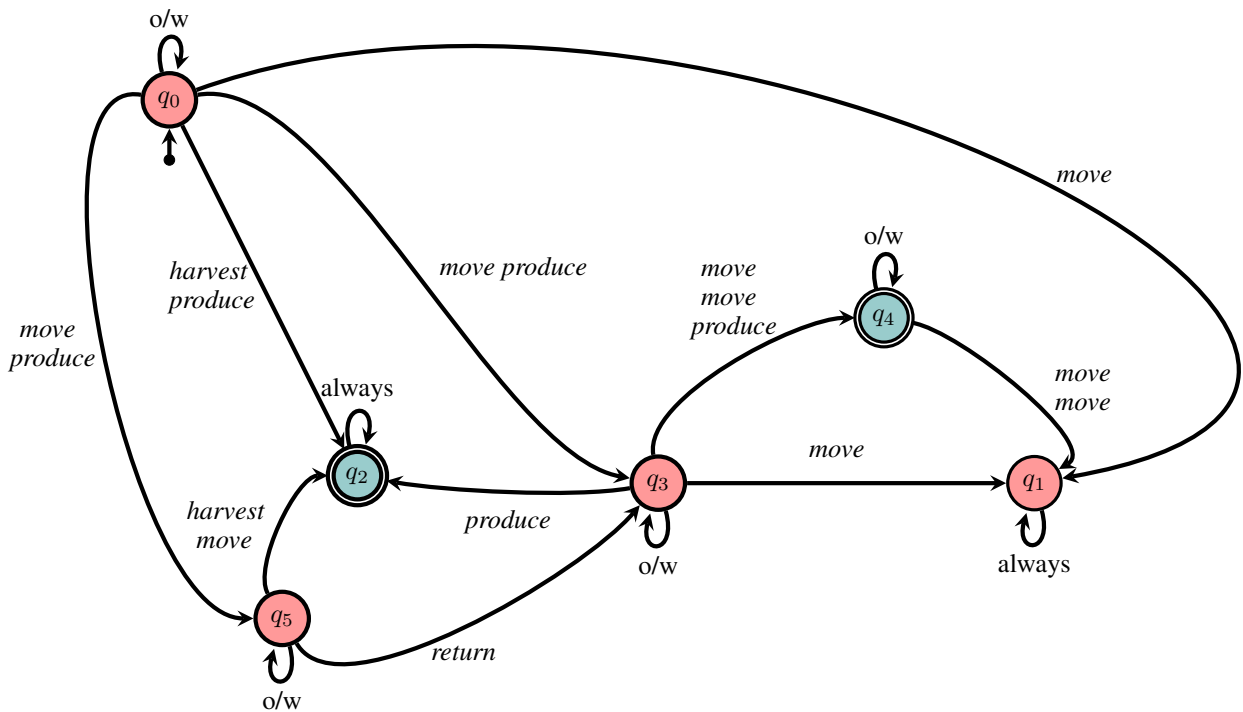
\begin{figure}[h!]
    \centering
    \vspace{-6mm}
    \begin{tikzpicture}[node distance=2cm,on grid,every initial by arrow/.style={ultra thick,->, >=stealth}, initial text={}]%
  \node[ultra thick,state,fill=red!40!white,minimum size=.5cm] (q_0) at (-8/2,4) {$q_0$};
  \node[ultra thick,state,fill=red!40!white,minimum size=.5cm]  (q_3) at (5/2,-1.1) {$q_3$};
  
  \node[very thick,state, fill=red!40!white,minimum size=.5cm]  (q_1) at (15/2,-1.1) {$q_1$};
  \node[ultra thick,state,accepting,fill=teal!40!white,minimum size=.5cm]  (q_2) at (-3/2,-1) {$q_2$};
  %\node[very thick,state,accepting, fill=teal!40!white,minimum size=.5cm]  (q_4) at (15/2,-1.4) {$q_4$};
  \node[very thick,state,accepting, fill=teal!40!white,minimum size=.5cm]  (q_4) at (11/2,1.1) {$q_4$};
  \node[ultra thick,state,fill=red!40!white,minimum size=.5cm]  (q_5) at (-5/2,-2.9) {$q_5$};

  % Initial node
  \path[ultra thick,->, >=stealth] (-8/2,3.2) edge node [right] {} (q_0);
  \draw[fill=black] (-8/2,3.2) circle (0.07);
  
  % Loops
  \path[->] (q_0) edge [ultra thick,->, >=stealth,loop above] node { \text{o/w}} ();
  \path[->] (q_1) edge [ultra thick,->, >=stealth,loop below] node { \text{always}} ();
  \path[->] (q_2) edge [ultra thick,->, >=stealth,loop above] node { \text{always}} ();
  \path[->] (q_3) edge [ultra thick,->, >=stealth,loop below] node { \text{o/w}} ();
  \path[->] (q_4) edge [ultra thick,->, >=stealth,loop above] node { \text{o/w}} ();
  \path[->] (q_5) edge [ultra thick,->, >=stealth,loop below] node { \text{o/w}} ();
  
  % Edges
  
  \path[ultra thick,->, >=stealth,text width=1cm] (q_0) edge node [left=0.1] { \textit{harvest \\ produce} } (q_2);
  
  \draw[ultra thick,->, >=stealth, out=10, in=170, looseness=0.6] (q_0) to node [right=0.2] {\text{{\textit{move produce}}}} (q_3);
  
  \draw[ultra thick,->, >=stealth, out=20, in=10, looseness=1.5] (q_0) to node [right=0.2] {\text{{\textit{move}}}} (q_1);
  
%   \draw[ultra thick,->, >=stealth, out=120, in=350, looseness=0.6] (q_2) to node [below=0.07] {\text{{\textit{writev}}}} (q_1);
  
%   \draw[ultra thick,->, >=stealth, out=190, in=-10, looseness=0.6] (q_3) edge node [below=0.1] {\text{{\textit{read}}}} (q_0);
  
  \draw[ultra thick,->, >=stealth, out=350, in=30, looseness=0.6] (q_4) edge node [right=0.2, text width=1cm, align=center] {{\textit{move\\move}}} (q_1);
  
  \path[ultra thick,->, >=stealth,text width=1cm] (q_3) edge node [above=0.2] { \textit{move} } (q_1);

  \draw[ultra thick,->, >=stealth, out=100, in=-180, looseness=0.6] (q_3) edge node [above=0.2,text width=1cm, align=center] {{\textit{move move\\produce}}} (q_4);
  
  \draw[ultra thick,->, >=stealth, out=170, in=-180, looseness=0.6] (q_0) edge node [left=0.1, text width=1cm, align=center] {{\textit{move\\produce}}} (q_5);
  
  \draw[ultra thick,->, >=stealth, out=120, in=-180, looseness=0.6] (q_5) edge node [left=0.1, text width=1cm, align=center] {{\textit{harvest \\ move}}} (q_2);
  
  \draw[ultra thick,->, >=stealth, out=300, in=-130, looseness=0.6] (q_5) edge node [below=0.1] {\text{{\textit{return}}}} (q_3);
  
  \draw[ultra thick,->, >=stealth, out=190, in=-10, looseness=0.6] (q_3) edge node [below=0.1] {\text{{\textit{produce}}}} (q_2);
  
%   \path[ultra thick,->, >=stealth] (q_3) edge node [right=0.2] {\textbf{\small A or B}} (q_1);
  
%   \path[ultra thick,->, >=stealth, out=2, in=190, looseness=0.8] (q_3) edge node [below] {\textit{{clock\_gettime}}} (q_2);
  
%   \path[ultra thick,->, >=stealth, out=90, in=-80, looseness=0.8] (q_2) edge node [above right] {\textit{getuid32}} (q_4);

\end{tikzpicture}%

% delta(0,3) = 1
%move 
% delta(0,104) = 2
%move_move_produce_produce_produce_return_return
% delta(0,116) = 2
%harvest_produce_produce_produce
% delta(0,134) = 2
%
% delta(0,138) = 2
% delta(0,141) = 2
% delta(0,4) = 2
% delta(0,64) = 2
% delta(0,5) = 3
% delta(0,2) = 5
% delta(3,3) = 1
% delta(3,1) = 2
% delta(3,6) = 2
% delta(3,4) = 4
% delta(4,6) = 1
% delta(5,48) = 2
% delta(5,11) = 3
% q_0 = 0
    \caption{A DFA learned from the StarCraft dataset by limiting the maximum number of states to 10. A decision is provided after each new observation based on the current state: yes for the blue accepting state, and no for the red, non-accepting states. “o/w” (otherwise) stands for all symbols that do not appear on outgoing edges from a state. “always” stands for all symbols.}
    \label{fig:supp_example_DFA_starcraft_10_state}
\end{figure}

\subsection{Linear Temporal Logic}

% LTL - syntax, semantics, notion of $\models$

% DECIDED AGAINST: Perhaps a bit about Chomsky Hierarchy, regular langauges, regular expressions, CFGs, 

%TODO - CAMERA READY 
% CITE OUR AAMAS18 work as well as others work.
%\subsection{Linear Temporal Logic}
%\commentms{usefulness of LTL \#1}\textcolor{red}{To this end, we propose the use of Linear Temporal Logic (LTL) as a compelling language for teaching multiple tasks to an RL agent in a manner that supports composition of learned skills. We also propose a novel algorithm that exploits LTL progression and off-policy RL to speed up learning without compromising convergence guarantees. Experiments over randomly generated Minecraft-like grids illustrate our superior performance relative to the state of the art.} \revisit{NEW CITATION} \cite{tor-etal-aamas18}

%TODO - CAMERA READY
% CITE OTHER LTL /LTLF PAPERS
%\LTL interpreted over \emph{finite} traces has received attention from the planning community \cite{pddl3,bai-mci-aaai06}. \acite{deg-var-ijcai13} provided a formal description and named it \LTLf.

% \subsection{Linear Temporal Logic}
In Section \ref{sec:interp} of our paper, we proposed Linear Temporal Logic (\LTL) as a candidate language for conveying explanations \emph{to} humans or other agents, and for use \emph{by} humans or other agents to express temporal properties that the agent might wish to add to the classifier or have verified. %LATER - against the data and/or classifier - think through a little more.
In what follows we review the basic syntax and semantics of \LTL \cite{pnueli1977temporal}. Note that \LTL formulae can be interpreted over either infinite or finite traces, with the finite interpretation requiring a small variation in the interpretation of formulae in the final state of the finite trace.  Here we describe \LTL interpreted over infinite traces noting differences as relevant. %is classically interpreted over an infinite sequence, or \emph{trace}, of states. \LTL interpreted over \emph{finite} traces has received attention from the planning community \cite{pddl3,bai-mci-aaai06}. \acite{deg-var-ijcai13} provided a formal description and named it \LTLf.

\LTL is a propositional logic language augmented with modal temporal  operators \emph{next} ($\ltlnext{}$) and \emph{until} ($\ltluntil{}{}$),
from which it is possible to define the well-known operators \emph{always} ($\ltlalways{}$), \emph{eventually} ($\ltleventually{}$), and \emph{release} ($\ltlrelease{}{}$).
When interpreted over finite traces, a 
\emph{weak next} ($\ltlweaknext{}$) operator is also utilized, and is equivalent to $\ltlnext{}$ when %iff 
$\pi$ is infinite. An \LTL formula over a set of propositions $\mathcal{P}$ is defined inductively: a proposition in $\mathcal{P}$ is a formula, and if $\psi$ and $\chi$ are formulae, then so are $\neg \psi$, $(\psi\wedge\chi)$, $(\ltluntil{\psi}{\chi})$,  $\ltlnext{\psi}$, and $\ltlweaknext{\psi}$.

%STOPPED HERE

The semantics of \LTL is defined as follows. A trace $\pi$ is a sequence of states, where each state is an element in $2^\mathcal{P}$.  We denote the first state of $\pi$ as $s_1$ and the $i$-th state of $\pi$ as $s_i$; $|\pi|$ is the length of $\pi$ (which is $\infty$ if $\pi$ is infinite). We say that $\pi$ satisfies $\varphi$ ($\pi\models\varphi$, for short) iff $\pi,1\models \varphi$, where for every %natural number 
$i\geq 1$:
\begin{itemize}
\item $\pi,i\models p$, for a propositional variable $p\in \mathcal{P}$, iff $p\in s_i$,
\item $\pi,i\models \neg \psi$ iff it is not the case that $\pi,i\models\psi$,
\item $\pi,i\models (\psi\wedge\chi)$ iff $\pi,i\models\psi$ and $\pi,i\models \chi$,
\item $\pi, i \models \ltlnext{} \varphi$ iff $i < |\pi|$ and $\pi, i+1 \models \varphi$, 
\item $\pi, i \models (\varphi_1 \ltluntil{}{} \varphi_2)$ iff for some $j$ in $\{i,\ldots,|\pi|\}$, it holds that $\pi, j \models \varphi_2$ and for all $k \in \{i,\ldots, j-1\}$, $\pi, k \models \varphi_1$,
\item $\pi, i \models \ltlweaknext{} \varphi$ iff $i = |\pi|$ or $\pi, i+1 \models \varphi$.
\end{itemize}
%Observe operator $\ltlweaknext{}$ is equivalent to $\ltlnext{}$ iff $\pi$ is infinite. As such we allow $\ltlweaknext{}{}$ in \LTL formulae, we do not use the acronym \LTLf, but we are explicit regarding which interpretation we use (either finite or infinite) when not obvious from the context. 
% {\color{red} As usual, $\ltleventually{\varphi}$ is defined as $(\ltluntil{\true}{\varphi})$, and $\ltlalways{\varphi}$ as $\neg \ltleventually{\neg \varphi}$. We use the \emph{release} ($\ltlrelease{}{}$) operator, defined by $(\ltlrelease{\psi}{\chi})\isdef \neg( \ltluntil{\neg \psi}{\neg \chi})$.}
$\ltleventually{\varphi}$ is defined as $(\ltluntil{\true}{\varphi})$,  $\ltlalways{\varphi}$ as $\neg \ltleventually{\neg \varphi}$, and $(\ltlrelease{\psi}{\chi})$ as  $\neg( \ltluntil{\neg \psi}{\neg \chi})$.

%\subsection{LTL, Automata, and Planning}
%Regardless of whether the interpretation is over an infinite or finite trace, 
Given an \LTL formula $\varphi$ there exists an automaton $\mathcal{A}_\varphi$ that accepts a trace $\pi$ iff $\pi\models\varphi$. It follows that, given a set of consistent \LTL formulae, $\{\varphi_1,\ldots,\varphi_n\}$, there exists an automaton, $\mathcal{A}_\varphi$,where $\varphi = \bigwedge_i \varphi_i$, that accepts a trace $\pi$
%SM The iff is wrong
%iff $\pi\models\varphi_i$, for all $i$. 
iff $\pi\models\varphi$.
As noted in Section \ref{sec:prelim} an automaton defines a language---a set of words that are accepted by the automaton.
We say that an automaton ${\cal A}$ satisfies an \LTL formula, $\varphi$, ${\cal A} \models \varphi$ iff for every accepting trace, $\pi_i$ of ${\cal A}$, $\pi_{i}\models\varphi$. Such satisfying \LTL formulae provide another means of explaining the behaviour of a DFA classifier. %SM redundant As noted in Section \ref{sec:prelim} the set of all words (traces) accepted by an automaton define a language.  
%Furthermore, given a set of consistent \LTL formulae, we can define an automaton that accepts traces that satisfy those formulae. 

Depending on whether \LTL formula, $\varphi$, is interpreted over finite or infinite traces, different types of automata are needed to capture $\varphi$. For the purposes of this paper, it is sufficient to know that DFAs are sufficiently expressive to capture any \LTL formulae interpreted over finite traces, but only a subset (a large and useful subset) of \LTL formulae interpreted over infinite traces.

\section{Experimental Evaluation}
\label{supp:exp}

\subsection{Experimental Setup}
\label{supp:exp:setup}
We first provide experimental details for each method used in our main set of experiments in Section \ref{sec:exp}. DISC, LSTM, and HMM used a validation set consisting of 20\% of the training traces per class on all domains except MIT-AR. This was since MIT-AR consisted of very limited training data, and using a validation set worsened performance in all cases. We describe the specific modifications for each method below. Additionally, minor changes were made for our experiments on multi-label classification (described in \ref{supp:multilabel}).

DISC (our approach) used Gurobi optimizer to solve the MILP formulation for learning DFAs. We set $q_\mathrm{max}$, the maximum possible number of states in a DFA, to 5 for Crystal Island and MIT-AR and 10 for all other domains along with a time limit of 15 minutes to learn each DFA. DISC also uses two regularization terms to prevent overfitting: a term penalizing the number of transitions between different states, with coefficient $\lambda_t$; and a term penalizing nodes not assigned to an absorbing state, with coefficient $\lambda_a$. We set $\lambda_a = 0.001$, and use a validation procedure to choose $\lambda_t$ from 11 approximately evenly-spaced values (on a logarithmic scale) between $0.0001$ and $10$, inclusive. The model with maximum $F_1$-score on the validation set is selected. For MIT-AR, instead of using a validation set, we choose $\lambda_t$ from a small set of evenly-spaced values ($\{3, 5.47, 10\}$) and select the model with highest training $F_1$-score. 

The DFA-FT baseline utilized the full tree of observations (rather than the prefix tree used in DISC) and learned one DFA per label. A single positive and negative DFA state were designated, and any node in the tree whose suffixes were all positive or negative were assigned to the positive or negative state, respectively. Every other node of the tree was assigned to a unique DFA state and attached with the empirical (training) probability of a trace being positive, given that it transitions through that DFA state. To classify a trace in the presence of multiple classes, all $|\mathcal{C}|$ DFAs were run in parallel, and the class of the DFA with highest probability was returned.

Our LSTM model consisted of two LSTM layers, a linear layer mapping the final hidden state to labels, and a log-softmax layer. The LSTM optimized a negative log-likelihood objective using Adam optimizer \cite{kingma2014adam}, with equal weight assigned to each prefix of the trace (to encourage early prediction). We observed inferior performance overall when using one or four LSTM layers. The batch size was selected from $\{8, 32\}$, the size of the hidden state from $\{25, 50 \}$, and the number of training epochs from $[1, 300]$ by choosing the model with the highest validation accuracy given full traces. For the MIT-AR dataset, the hyperparameters were hand-tuned to 8 for batch size, 25 for hidden dimension, and 75 epoches.

Our HMM model was based on an open-source Python implementation for unsupervised HMMs from Pomegranate\footnote{\url{https://pomegranate.readthedocs.io/en/latest/}}. We trained a separate HMM for each class, and classify a trace by choosing the HMM with highest probability. Each HMM was trained with the Baum-Welch algorithm using a stopping threshold of $0.001$ and a maximum of $10^6$ iterations. The validation set was used to select the number of discrete hidden states from $\{5, 10\}$ and a pseudocount (for smoothing) from $\{0, 0.1, 1\}$. For MIT-AR we hand-tuned these hyperparameters to $10$ for the number of hidden states and $1$ for the pseudocount. 

The n-gram models did not require validation. We used a smoothing constant $\alpha = 0.5$ to prevent estimating a probability of 0 for unseen sequences of observations.

\subsection{Datasets}
\label{supp:datasets}

The StarCraft and Crystal Island datasets were obtained thanks to the authors \cite{ha2011goal,kantharajuMCTS2019}, while the malware datasets, ALFRED, and MIT-AR are publicly available\footnote{\url{https://github.com/mlbresearch/syscall-traces-dataset}}\footnote{\url{https://github.com/askforalfred/alfred/tree/master/data}}\footnote{\url{https://courses.media.mit.edu/2004fall/mas622j/04.projects/home/}}.

\subsubsection*{Malware}
The two malware datasets (BootCompleted, BatteryLow) were generated by \citeauthor{bernardi2019dynamic} (\citeyear{bernardi2019dynamic}) by downloading and installing various malware applications with various intents (e.g., wiretapping, selling user information, advertisement, spam, stealing user credentials, ransom) on an Android phone. Each dataset reflects an Android operating system event (e.g., the phone's battery is at 50\%) that is broadcasted system-wide (such that the broadcast also reaches every active application, including the running malware). Each family of malware is designed to react to a system event in a certain way, which can help distinguish it from the other families of malware (see Table 4 in \cite{bernardi2019dynamic} for the list of malware families used in the dataset). 

A single trace in the dataset comprises a sequence of `actions' performed by the malware application (e.g., the system call \textit{clock\_gettime}) in response to the Android system call in question, and labelled with the class label corresponding to the particular malware family.

\subsubsection*{StarCraft}

The StarCraft dataset was constructed by \citeauthor{kantharajuMCTS2019} (\citeyear{kantharajuMCTS2019}) by using replay data of StarCraft games where various scripted agents were playing against one another. To this end, the real-time strategy testbed MicroRTS\footnote{\url{https://github.com/santiontanon/microrts}} was used. The scripted agents played in a 5-iterations round-robin tournament with the following agent types: \textit{POLightRush, POHeavyRush, PORangedRush, POWorkerRush, EconomyMilitaryRush, EconomyRush, HeavyDefense, LightDefense, RangedDefense, WorkerDefense, WorkerRushPlusPlus}. Each agent competed against all other agents on various maps. 

A replay for a particular game comprises a sequence of both players' actions, from which the authors extracted one labelled trace for each player. 
We label each trace with the agent type (e.g. \textit{WorkerRushPlusPlus}) that generated the behaviour.

\subsubsection*{Crystal Island}
Crystal Island is an educational adventure game designed for middle-school science students \cite{ha2011goal, min2016player}, with the dataset comprising in-game action sequences logged from students playing the game. 
\textit{``In Crystal Island, players are assigned a single high-level objective: solve a science mystery. Players interleave periods of exploration and deliberate problem solving in order to identify a spreading illness that is afflicting residents on the island. In this setting, goal recognition involves predicting the next narrative sub-goal that the player will complete as part of investigating the mystery"} \cite{ha2011goal}. Crystal Island is a particularly challenging dataset due to players interleaving exploration and problem solving which leads to noisy observation sequences.

A single trace in the dataset comprises a sequence of player actions, labelled with a single narrative sub-goal (e.g., \textit{speaking with the camp’s virus expert} and see Table 2 in \cite{ha2011goal}). Each observation in the trace includes one of 19 player action-types (e.g., testing an object using the laboratory’s testing equipment) and one of 39 player locations. Each unique pair of action-type and location is treated as a distinct observation token. 

\subsubsection*{ALFRED}
ALFRED (Action Learning From Realistic Environments and Directives) is a benchmark for learning a mapping from natural language instructions and egocentric vision to sequences of actions for household tasks. We generate our training set from the set of expert demonstrations in the ALFRED dataset which were produced by a classical planner given the high-level environment dynamics, encoded in PDDL \cite{mcdermott1998pddl}. Task-specific PDDL goal conditions (e.g., rinsing off a mug and placing it in the coffee maker) were then specified and given to the planner, which generated sequences of actions (plans) to achieve these goals. There are 7 different task types which we cast as the set of class labels $\classLabels$ (see Figure 2 in \cite{ALFRED20}): \textit{Pick \& Place; Stack \& Place; Pick Two \& Place; Examine in Light; Heat \& Place; Cool \& Place; Clean \& Place}.

A single trace in the dataset comprises a sequence of actions taken by the agent in the virtual home environment, labelled with one of the class labels described above (e.g., \textit{Heat \& Place}).

\subsubsection*{MIT Activity Recognition (MIT-AR)}
MIT-AR was generated by \citeauthor{tapia2004activity} (\citeyear{tapia2004activity}) by collecting sensor data over a two week period from multiple sensors installed in a myriad of everyday objects such as drawers, refrigerators and containers. The sensors recorded opening and closing events related to these objects while the human subject carried out everyday activities. The resulting noisy sensor sequence data was manually labelled with various high-level daily activities in which the human subject was engaged. The activities in this dataset (which serve as the class labels in our experiments) include \textit{preparing dinner}, \textit{listening to music}, \textit{taking medication} and \textit{washing dishes}, and are listed in Table 5.3 in \cite{tapia2004activity}. In total there are 14 activities.

A single trace in the dataset comprises a sequence of sensor recordings (e.g., \textit{kitchen drawer interacted with} or \textit{kitchen washing machine interacted with}), labelled with one of the class labels described above (e.g., \textit{washing dishes}).

\subsection{Additional Results}

We display the extensive results from the main paper in Figures \ref{fig:r1}, \ref{fig:r2}, \ref{fig:r3}. For each domain, we present a line plot displaying the Cumulative Convergence Accuracy (CCA) up to the maximum length of any trace and a bar plot displaying the PCA at 20\%, 40\%, 60\%, 80\%, and 100\% of observations. Error bars report a 90\% confidence interval over 30 runs. 

\begin{figure}
    \centering
    
    Crystal Island
    
    \includegraphics[width=\textwidth]{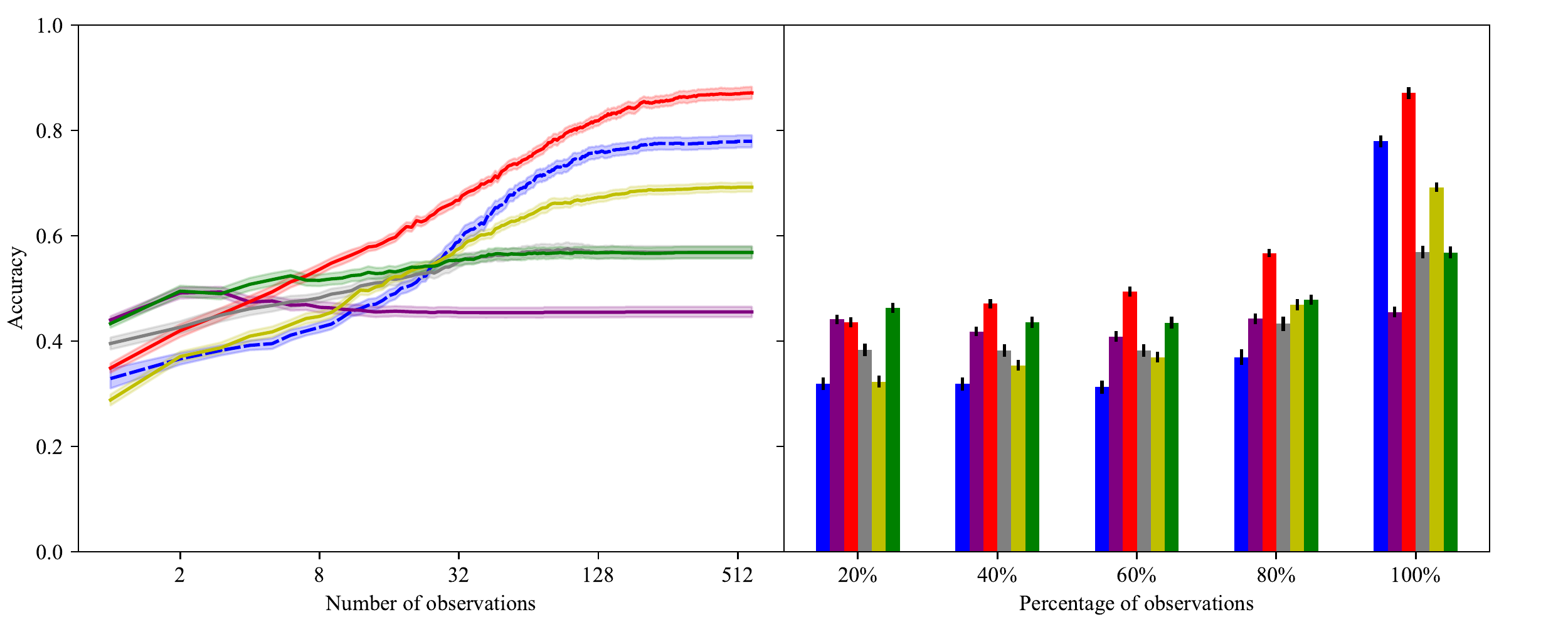}
    
    StarCraft
    
    \centering
    \includegraphics[width=\textwidth]{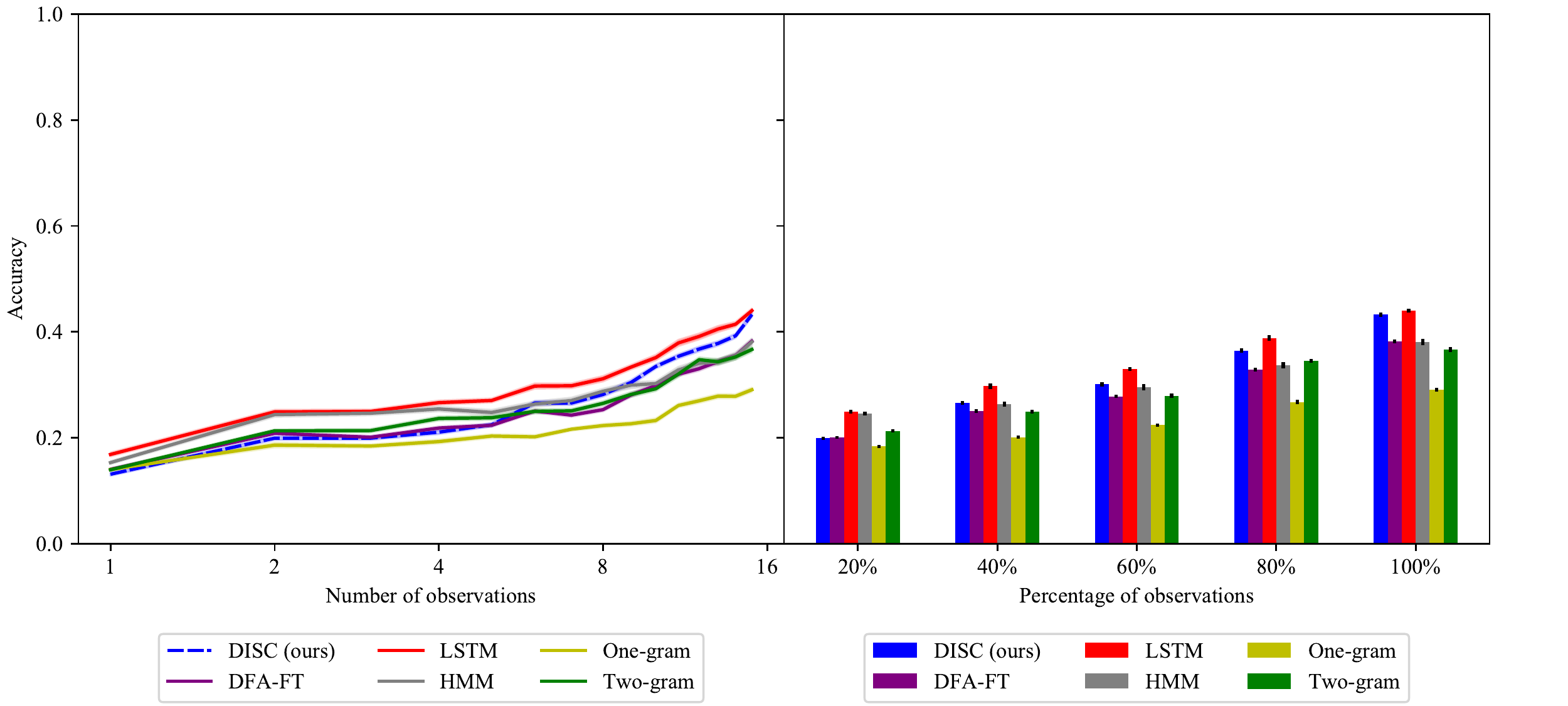}
    
    \caption{Results for the Crystal Island and StarCraft domains.}
    \label{fig:r1}
\end{figure}

\begin{figure}
    \centering
    ALFRED
    
    \includegraphics[width=\textwidth]{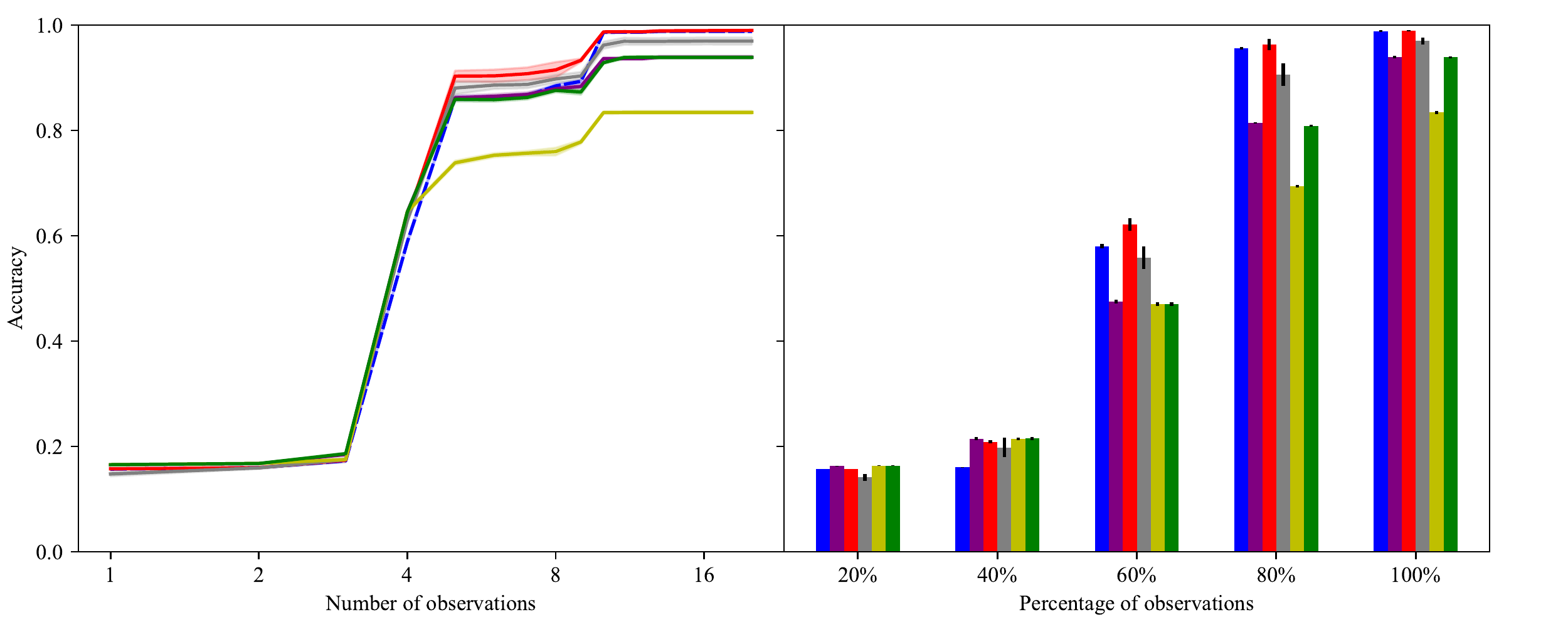}
    
    MIT-AR
    \includegraphics[width=\textwidth]{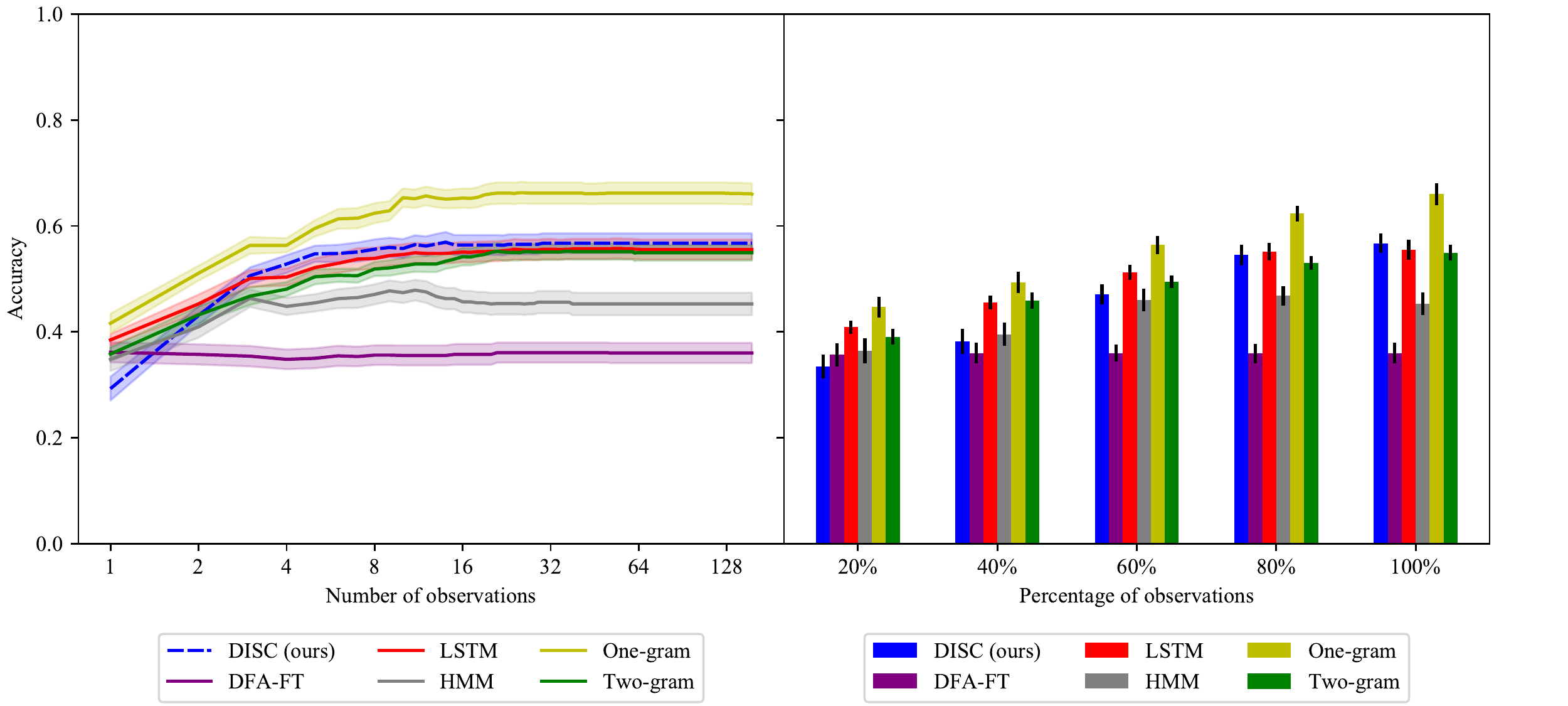}
    
    \caption{Results for the Alfred and MIT-AR domains.}
    \label{fig:r2}
\end{figure}

\begin{figure}    
     \centering
    BootCompleted
    
    \includegraphics[width=\textwidth]{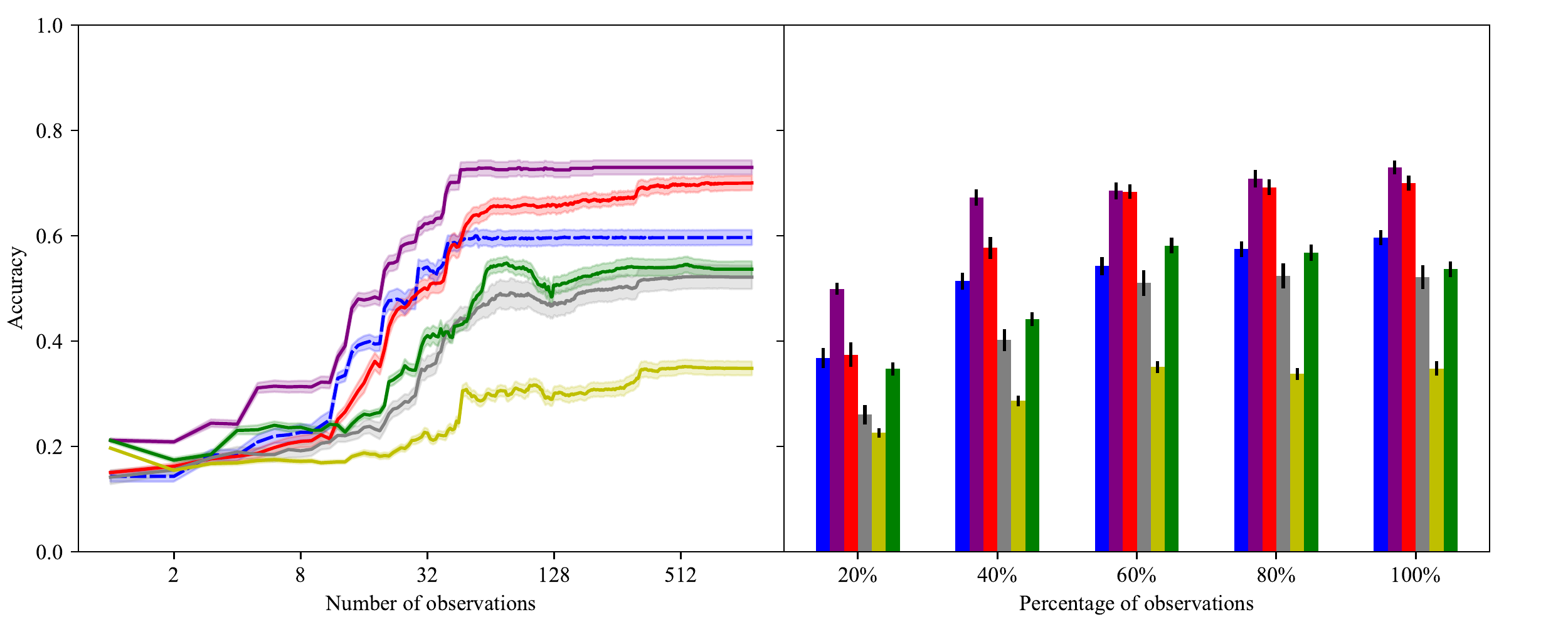}
    
    BatteryLow
    
    \includegraphics[width=\textwidth]{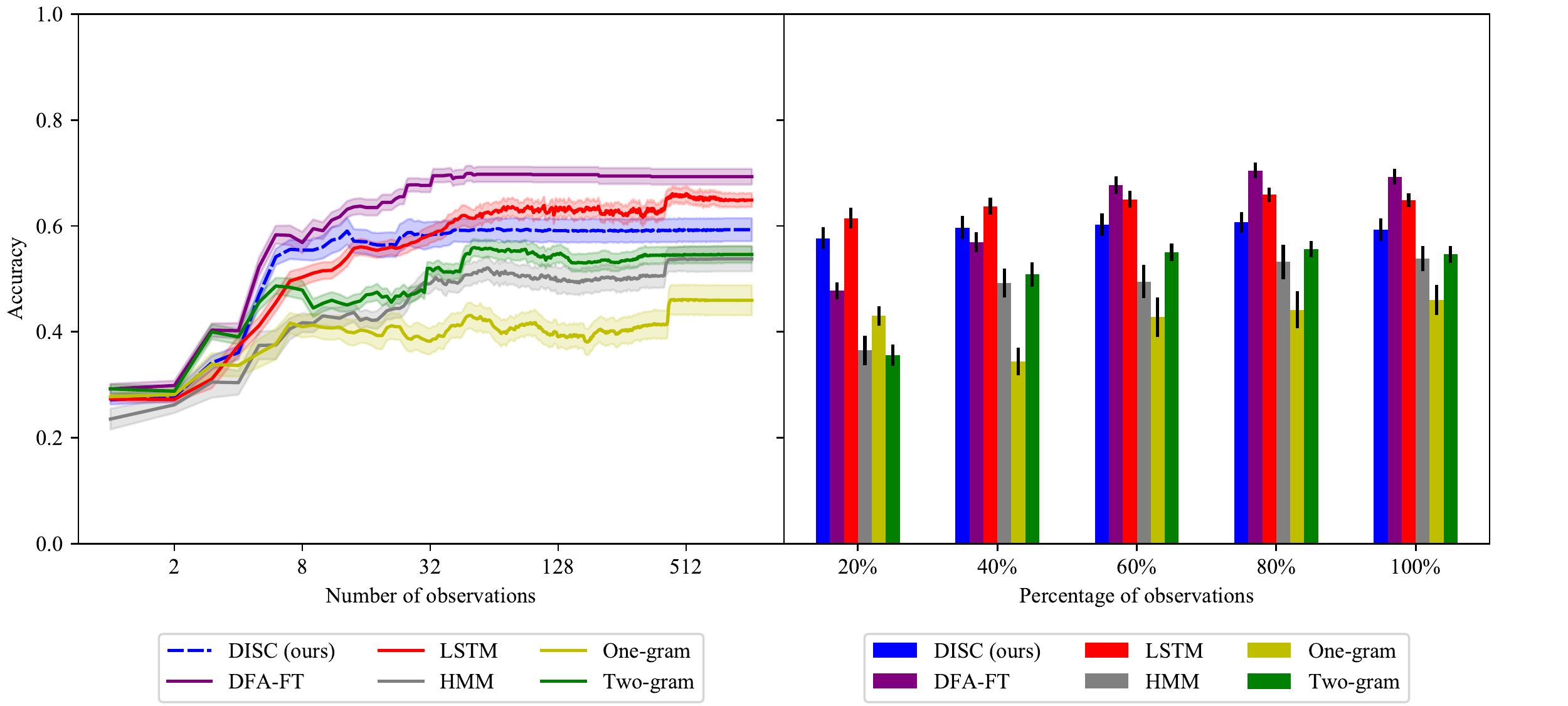}
    
    \caption{Results for the BootCompleted and BatteryLow domains.}
    \label{fig:r3}
\end{figure}

We further report in Table \ref{tab:dfa-size} the average number of states and transitions in learned DFAs for DISC and DFA-FT. DFAs learned using DISC were generally an order of magnitude smaller than DFAs learned using DFA-FT. 

\begin{table*}
\centering
\begin{tabular}{l|p{1.5cm}p{3cm}}
                        & \multicolumn{2}{c}{(\# DFA states, \# state transitions)} \\ \hline
Dataset                 &  \textbf{DISC} & DFA-FT \\ \hline

StarCraft               & 8.8, 37.2 & 170.4, 196.0    \\
MIT-AR                  & 3.0, 1.83  & 18.3, 103.4     \\
Crystal Island          & 3.9, 26.2 & 166.6, 451.4    \\
ALFRED                  & 5.1, 9.0 & 26.8, 53.2     \\
BootCompleted           & 9.7, 42.7 & 321.3, 391.5  \\
BatteryLow              & 8.9, 27.2 & 325.1, 365.7   \\
\end{tabular}
\caption{ The average number of DFA states (first), and the average number of state transitions (second) in learned models for \textbf{DISC} (ours) and DFA-FT over twenty runs, using the experimental procedure in Section \ref{supp:exp:setup}. }
\label{tab:dfa-size}
\end{table*}

\subsection{Early Classification}
\label{supp:early}

The two key problems in early prediction are: (1) to maximize accuracy given only a prefix of the sequence and (2) to determine a stopping rule for when to make a classification decision. (1) is not significantly different from vanilla sequence classification, thus, most work in early prediction focuses on (2). While many different stopping rules have been proposed in the literature, the correct choice should be task-dependent as it requires making a trade-off between accuracy and earliness. Furthermore, it is difficult to objectively compare early prediction models that may make decisions at different times. Our early classification experiment is designed to evaluate two essential criteria: the accuracy of early classification, and the accuracy of classifier confidence, while remaining independent of choice of stopping rule. 

Thus, we expand upon the early classification setting briefly mentioned in Section \ref{sec:results} where an agent can make an irrevocable classification decision after any number of observations, but prefers to make a correct decision as early as possible. This is captured by a non-increasing utility function $U(t)$ for a correct classification. Note the agent can usually improve its chance of a correct prediction by waiting to see more observations. If the agent's predictive accuracy after $t$ observations is $p(t)$, then to maximize expected utility, the agent should make a decision at time $t^* = \mathrm{argmax}_t \{ U(t)p(t) \}$. However, the agent only has access to its estimated confidence measures $\mathrm{conf}(t) \approx p(t)$. Thus, success in this task requires not only high classification accuracy, but also accurate confidence in one's own predictions. 

We test this setting on a subset of domains, with utility function $U(t) = \max \{ 1 - \frac{t}{40}, 0 \}$. We make the assumption that at time $t$, the classifier only has access to the first $t$ observations, but has full access to the values of $\mathrm{conf}(t')$ for all $t'$ and can therefore choose the optimal decision time. We only consider baselines which produce a probability distribution over labels (DISC, LSTM, n-gram), defining the classifier's confidence to be the probability assigned to the predicted label (i.e. the most probable goal).

Results are shown in Table \ref{tab:early}. DISC has a strong performance on each domain, only comparable by LSTM. This suggests the confidence produced by DISC accurately approximates its true predictive accuracy.

\begin{table}[tb]
\centering
\begin{tabular}{l|cccc}
          & \multicolumn{4}{c}{Average utility}                                    \\ \hline
Dataset   & \textbf{DISC}       & LSTM     & 1-gram          & 2-gram          \\ \hline
    ALFRED      & $0.840 (\pm 0.014)$           & {$\mathbf{0.855} (\pm 0.003)$}   & $0.703 (\pm 0.002)$  & $0.792 (\pm 0.003)$  \\
    
StarCraft     & $0.337 (\pm 0.005)$           & $\mathbf{0.341} (\pm 0.005)$   & $0.194 (\pm 0.004)$  & $0.273 (\pm 0.006)$  \\ 

BootCompleted              & $0.203 (\pm 0.014)$           & $\mathbf{0.218} (\pm 0.018)$   & $0.113 (\pm 0.003)$ & $0.157 (\pm 0.006)$
\end{tabular}
\caption{Results for the early classification experiment. Average utility per trace over twenty runs is reported with 90\% confidence error, with the best mean performance in each row in bold.}
\label{tab:early}
\end{table}

\subsection{Multi-label Classification}
\label{supp:multilabel}

In the goal recognition datasets used in our work we assume agents pursue a 
single goal achieved by the sequence of actions encoded in the sequence of observations. However, often times an agent will pursue multiple goals concurrently, interleaving actions such that each action in a trace is aimed at achieving any one of multiple goals. For instance, if an agent is trying to make toast \textit{and} coffee, the first action in their plan may be to fill the kettle with water, the second action may be to put the kettle on the stove, their third action might be to take bread out of the cabinet, and so on. We cast this generalization of the goal recognition task as a multi-label classification problem where each trace may have one or more class labels (e.g., \textit{toast} and \textit{coffee}).

We experiment with a synthetic kitchen dataset \cite{harman2020action} where an agent is pursuing multiple goals and non-deterministically switching between plans to achieve them. A single trace in this dataset comprises actions performed by the agent in the kitchen environment in pursuit of multiple goals drawn from the set of possible goals. Each trace is labelled with multiple class labels corresponding to the goals achieved by the interleaved plans encoded in the trace. In total there are 7 goals the agent may be pursuing ($|\classLabels| = 7$) and 25 unique observations ($|\Sigma| = 25$). The multi-goal kitchen dataset was obtained with thanks to the authors \cite{harman2020action}.

We modify DISC for this setting by directly using the independent outputs of the binary one-vs-rest classifiers---Bayesian inference is no longer necessary since we do not need to discriminate a single label. Precisely, for a given trace $\tau$, we independently run all $|\mathcal{C}|$ DFA classifiers and return \emph{all} classes for which the corresponding DFA accepts. We also disable the reweighting technique described in \ref{supp:prefix_tree_to_dfa} (i.e. by setting $\lambda^+ = \lambda^- = 1$) to focus on optimizing accuracy. We set DISC's hyperparameters to $q_\mathrm{max} = 5, \lambda_a = 0.001, \lambda_t = 0.0003$. The LSTM baseline is modified to return a $|\mathcal{C}|$-dimensional output vector containing an independent probability for each class and is trained with a cross-entropy loss averaged over all dimensions. At test time, we predict all classes $c \in \mathcal{C}$ with probability greater than 0.5. We set the LSTM's hyperparameters to 8 for batch size, 25 for hidden dimension size, and 250 for number of epoches. Results are shown in Figure \ref{fig:kitchen}, where we report the mean accuracy, averaged over all goals, over 30 runs. DISC achieves similar performance to LSTM (c) on this task. 

\begin{figure}
    \centering
    \includegraphics[width=\textwidth]{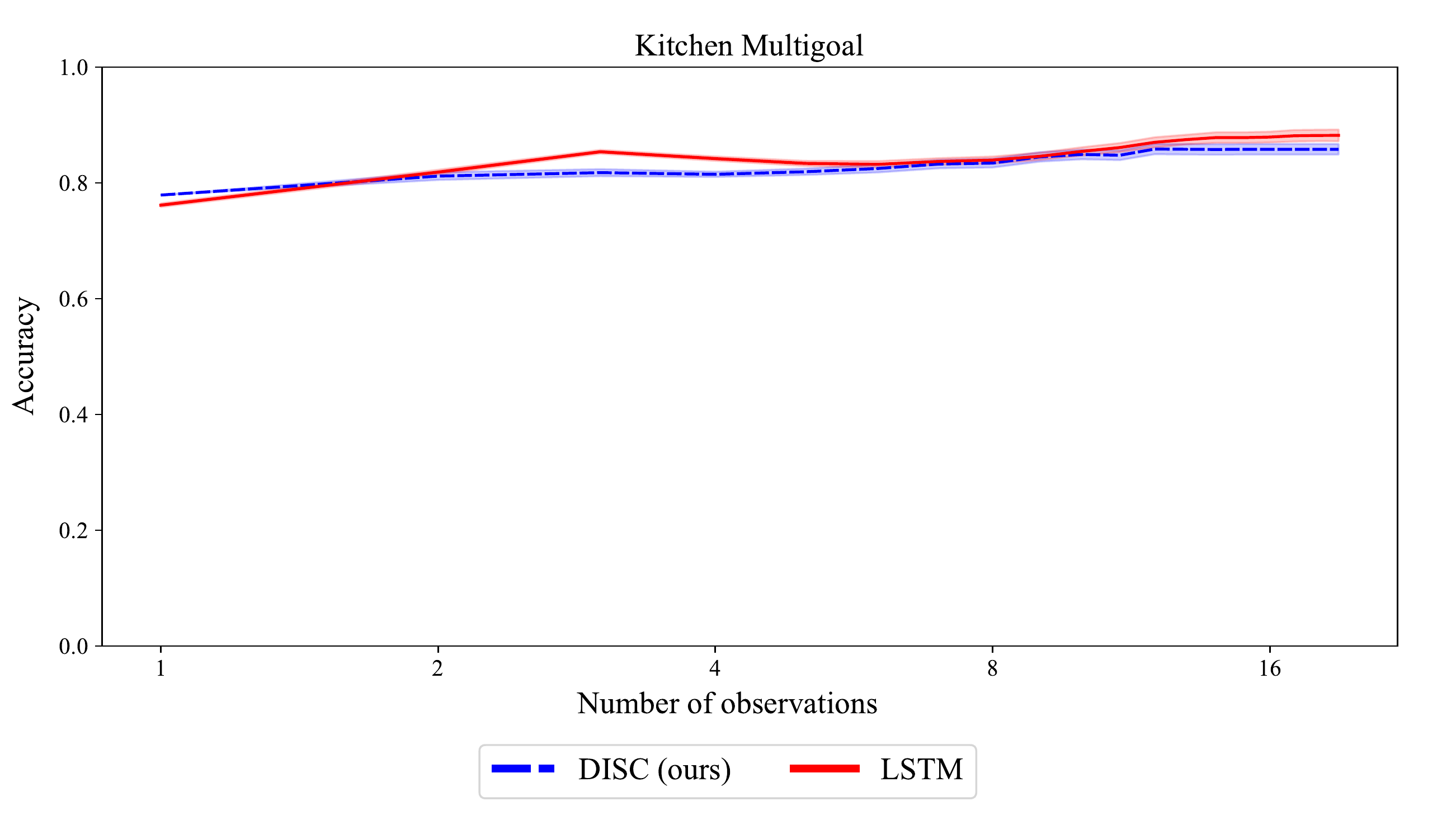}

    \caption{CCA for the Kitchen domain. Error bars represent a 90\% confidence interval over 30  runs. }
    \label{fig:kitchen}
\end{figure}

\end{document}